\documentclass[10pt,twocolumn,letterpaper]{article}

\PassOptionsToPackage{numbers, sort&compress}{natbib}

\usepackage[pagenumbers]{cvpr} %

\usepackage[utf8]{inputenc} %
\usepackage[T1]{fontenc}    %
\usepackage{url}            %
\usepackage{booktabs}       %
\usepackage{amsfonts}       %
\usepackage{nicefrac}       %
\usepackage{microtype}      %

\usepackage[table,x11names,usenames,dvipsnames]{xcolor}
\usepackage{tikz}
\usepackage[framemethod=tikz]{mdframed}
\usepackage{microtype}
\usepackage{graphicx}
\usepackage[export]{adjustbox} %
\usepackage{booktabs} %

\usepackage{amsmath}
\usepackage{amssymb}
\usepackage{mathtools}
\usepackage{amsthm}
\usepackage{stmaryrd}

\usepackage{tabularx}
\usepackage{makecell}

\usepackage{url}            %
\usepackage{booktabs}       %
\usepackage{amsfonts}       %
\usepackage{nicefrac}       %
\usepackage{microtype}      %
\usepackage{multirow}       %
\usepackage{algorithm}
\usepackage{algorithmic}

\usepackage{wrapfig}
\usepackage{comment}
\usepackage{amsmath,amssymb} %
\usepackage{symbols}
\usepackage{multirow}
\usepackage{pifont}%

\usepackage{amsmath,amsfonts,bm}

\def\1{\bm{1}}

\def\sp{space}

\def\rvc{{\mathbf{c}}}
\def\rvd{{\mathbf{d}}}

\def\rvo{{\mathbf{o}}}

\def\rvz{{\mathbf{z}}}

\DeclareMathAlphabet{\mathsfit}{\encodingdefault}{\sfdefault}{m}{sl}
\SetMathAlphabet{\mathsfit}{bold}{\encodingdefault}{\sfdefault}{bx}{n}

\newcommand{\E}{\mathbb{E}}

\newcommand*{\ShowNotes}{} %
\usepackage{color}
\usepackage{soul}
\usepackage{multirow}
\usepackage{xcolor}
\usepackage{wrapfig}

\definecolor{darkred}{rgb}{0.7,0.1,0.1}
\definecolor{darkgreen}{rgb}{0.1,0.7,0.1}
\definecolor{cyan}{rgb}{0.7,0.0,0.7}
\definecolor{dblue}{rgb}{0.2,0.2,0.8}
\definecolor{maroon}{rgb}{0.76,.13,.28}
\definecolor{burntorange}{rgb}{0.81,.33,0}
\definecolor{tealblue}{rgb}{0.212,0.459, 0.533}
\definecolor{myyellow}{rgb}{0.8627451 , 0.75294118, 0.20784314]}

\definecolor{mypink}{rgb}{0.93359375, 0.62109375, 0.83984375}

\definecolor{pp}{rgb}{0.43921569, 0.18823529, 0.62745098}
\definecolor{rr}{rgb}{0.5254902 , 0.00784314, 0.12941176}
\definecolor{bb}{rgb}{0.09019608, 0.23529412, 0.37647059}
\definecolor{yy}{rgb}{0.49803922, 0.3372549 , 0.0}
\definecolor{gg}{rgb}{0.02352941, 0.3372549 , 0.17647059}

\ifdefined\ShowNotes
  \newcommand{\colornote}[3]{{\color{#1}\bf{#2: #3}\normalfont}}
\else
  \newcommand{\colornote}[3]{}
\fi

\definecolor{lightred}{rgb}{0.9,0.4,0.4}

\newcommand{\eat}[1]{} %

\newlength\savewidth

\definecolor{turquoise}{cmyk}{0.65,0,0.1,0.1}
\definecolor{purple}{rgb}{0.65,0,0.65}
\definecolor{darkgreen}{rgb}{0.0, 0.5, 0.0}
\definecolor{darkred}{rgb}{0.5, 0.0, 0.0}
\definecolor{darkblue}{rgb}{0.0, 0.0, 0.5}
\definecolor{blue}{rgb}{0.0, 0.0, 1.0}
\definecolor{orange}{rgb}{1.0,0.5,0.0}

\newcommand{\hide}[1]{{}}

\makeatletter
\renewcommand{\paragraph}{%
  \@startsection{paragraph}{4}%
  {\z@}{0.3ex \@plus 1ex \@minus .1ex}{-1em}%
  {\normalfont\normalsize\bfseries}%
}
\makeatother

\newif\ifproofread

\usepackage[multiple]{footmisc}

\definecolor{mybrown}{rgb}{0.87058824, 0.56078431, 0.01960784}
\definecolor{myblue}{rgb}{0.3372549 , 0.70588235, 0.91372549}
\definecolor{mypurple}{rgb}{0.8, 0.47058824, 0.7372549 }
\definecolor{myorange}{rgb}{0.835, 0.368, 0}
\definecolor{mygreen}{rgb}{0.00784314, 0.61960784, 0.45098039}
\definecolor{mygt}{rgb}{0.0078125 , 0.57421875, 0.40625}
\definecolor{mysp}{rgb}{0.84765625, 0.515625  , 0.0234375}
\definecolor{mycitecolor}{rgb}{0,0.08,0.45}
\definecolor{mygr}{rgb}{0.9607,0.9607,0.9607}
\definecolor{myoo}{rgb}{0.992,0.9176,0.9019}

\definecolor{myrr}{HTML}{AE031A}
\definecolor{mybb}{HTML}{0155B3}

\usepackage{pifont}
\usepackage[T1]{fontenc}
\usepackage[utf8]{inputenc}
\usepackage{pmboxdraw}
\usepackage{thm-restate}
\definecolor{cvprblue}{rgb}{0.21,0.49,0.74}
\usepackage[pagebackref,breaklinks,colorlinks,citecolor=cvprblue]{hyperref}
\usepackage{url}            %
\usepackage{booktabs}       %
\usepackage{amsfonts}       %
\usepackage{nicefrac}       %
\usepackage{microtype}      %
\usepackage{multirow}       %
\usepackage{algorithm}
\usepackage{algorithmic}
\usepackage{mathtools}%

\mdfdefinestyle{MyFrame}{%
    linecolor=Black,
    outerlinewidth=0.1pt,
    roundcorner=2pt,
    innertopmargin=0.05\baselineskip,
    innerbottommargin=0.3\baselineskip,
    innermargin = 0.1cm,
    outermargin = 0.cm,
    innerrightmargin=5pt,
    innerleftmargin=5pt,
    backgroundcolor=Gray!10!white}

\usepackage{arydshln}

\usepackage{listings}
\usepackage{symbols}

\newcommand{\myparagraph}[1]{\vspace*{1pt}{\bf\noindent #1}}

\def\paperID{3352}
\def\confName{CVPR}
\def\confYear{2024}

\title{
Alpha Invariance: On Inverse Scaling Between Distance and Volume Density in Neural Radiance Fields
}

\author{Joshua Ahn$^{*}$\\
University of Chicago\\
{\tt\small jjahn@uchicago.edu}
\and
Haochen Wang$^{*}$\\
TTI-Chicago\\
{\tt\small whc@ttic.edu}
\and
Raymond A. Yeh\\
Purdue University\\
{\tt\small rayyeh@purdue.edu}
\and
Greg Shakhnarovich\\
TTI-Chicago\\
{\tt\small greg@ttic.edu}
}

\begin{document}

\twocolumn[{
\renewcommand\twocolumn[1][]{#1}
\maketitle
}]

\begin{abstract}
Scale-ambiguity in 3D scene dimensions leads to magnitude-ambiguity of volumetric densities in neural radiance fields, i.e., the densities double when scene size is halved, and vice versa. We call this property alpha invariance. For NeRFs to better maintain alpha invariance, we recommend 1) parameterizing both distance and volume densities in log space, and 2) a discretization-agnostic initialization strategy to guarantee high ray transmittance. We revisit a few popular radiance field models and find that these systems use various heuristics to deal with issues arising from scene scaling. We test their behaviors and show our recipe to be more robust.
Visit our project page at \href{https://pals.ttic.edu/p/alpha-invariance}{https://pals.ttic.edu/p/alpha-invariance}.
\end{abstract}

{
\let\thefootnote\relax
\footnote{* Equal contribution.}
}

\section{Introduction}

3D computer graphics and vision are fundamentally scale-ambiguous. Lengths of objects and scenes are often unitless and measure up to a constant ratio. This is fine for routines such as projection, triangulation, and camera motion estimation. A common practice is to normalize the scene dimension or the size of a reference object to an arbitrary number.  %

Volume rendering~\cite{maxvol} used by Neural Radiance Fields (NeRFs)~\cite{nerf} presents a complication due to explicit integration over space. Since the distance scaling is arbitrary, %
the volumetric density function $\sigma(x)$ must compensate for the multiplicative factor to render the same final RGB color. In other words, if the scene size expands by a factor $k$, it is \emph{sufficient} for the learned $\sigma$ to shrink by $1/k$ to be invariant to the change. The same applies to integration over cones or more general spatial data structures~\cite{mipnerf, zipnerf}. Note that in addition to scene dimensions, the magnitude of $\sigma$ is also influenced by the number of samples per ray and ultimately the sharpness of change in local opacity. %

\begin{figure}[t]
\centering
\vspace{-0.5cm}
\includegraphics[width=0.85\linewidth]{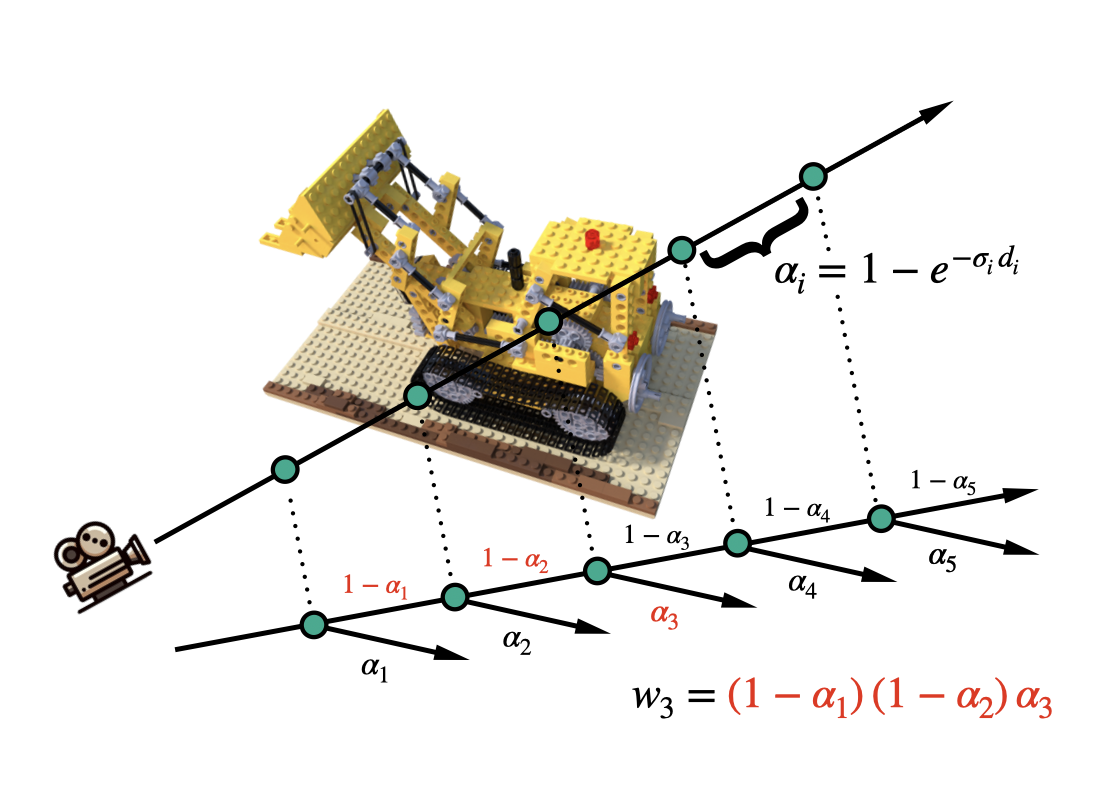}
\vspace{-0.6cm}
\caption{
     A discretized view of volume rendering. \textbf{Top}: a ray is cut into intervals, each with a density $ \sigma_i \ge 0 $ and interval length $ d_i $. \textbf{Bottom:} illustration of the weight given to the 3rd interval, computed through alpha compositing. The rendered color is obtained by weighting all the interval colors with their $w_i$s.
     If we scale each $d_i$ by a constant $k$, scaling $\sigma_i$ by $\frac{1}{k}$ renders the identical color.
}
\vspace{-0.5cm}
\label{fig:volrend}
\end{figure}

There is no single ``correct'' size for a scene setup. A robust algorithm should be able to perform consistently across different scalings. We investigate how this
notion of invariance manifests itself in practice,
discuss how the hyperparameter decisions affect it, and propose a solution that ensures robustness to distance scaling across the NeRF methods.
We revisit and experiment with a few popular NeRF architectures:
Vanilla NeRF~\cite{nerf},
TensoRF~\cite{tensorf},
DVGO~\cite{dvgo},
Plenoxels~\cite{plenoxels},
and
Nerfacto~\cite{nerfstudio}, and find that many systems use tailored heuristics that work well at a particular scene size. We analyze the impact of some critical hyperparameters which are often overlooked or not %
highlighted
in the original papers.

\begin{figure*}[!tbh]
\centering
\begin{tabular}{ccc}
\includegraphics[height=3.93cm]{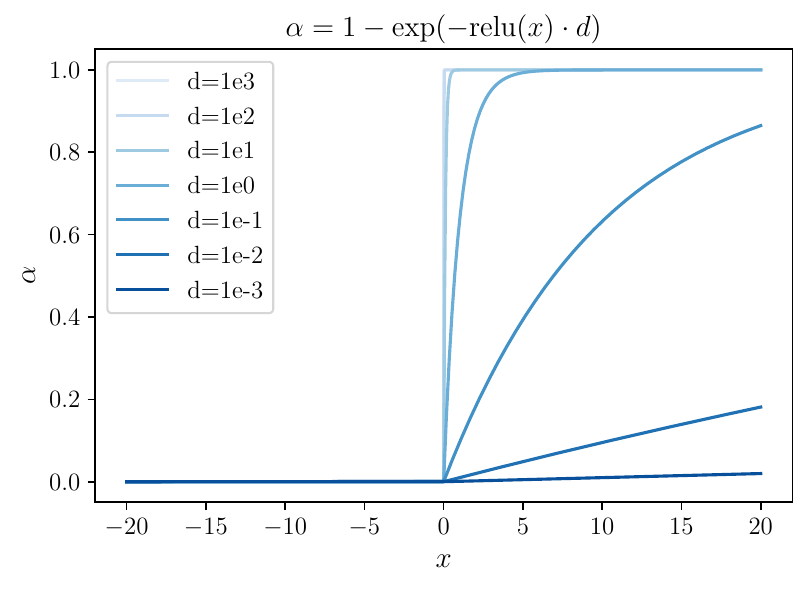}&
\includegraphics[height=3.93cm]{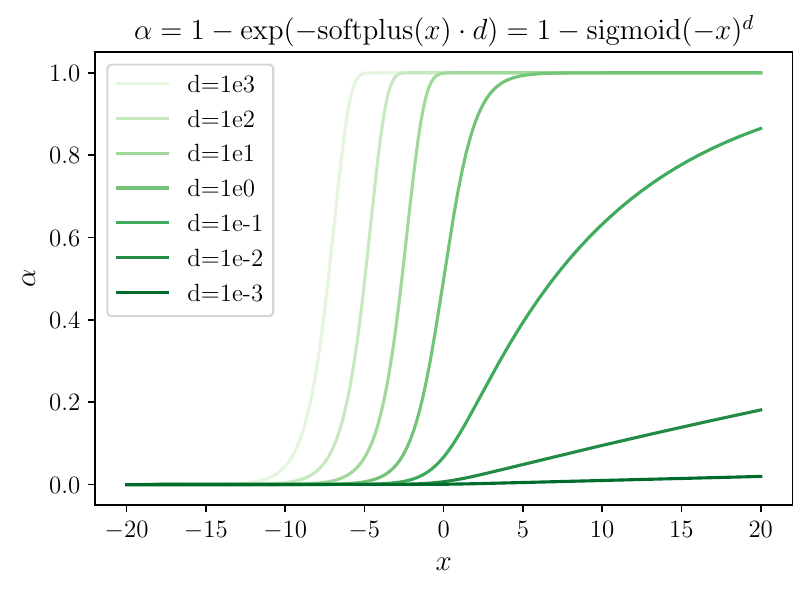}&
\includegraphics[height=3.93cm]{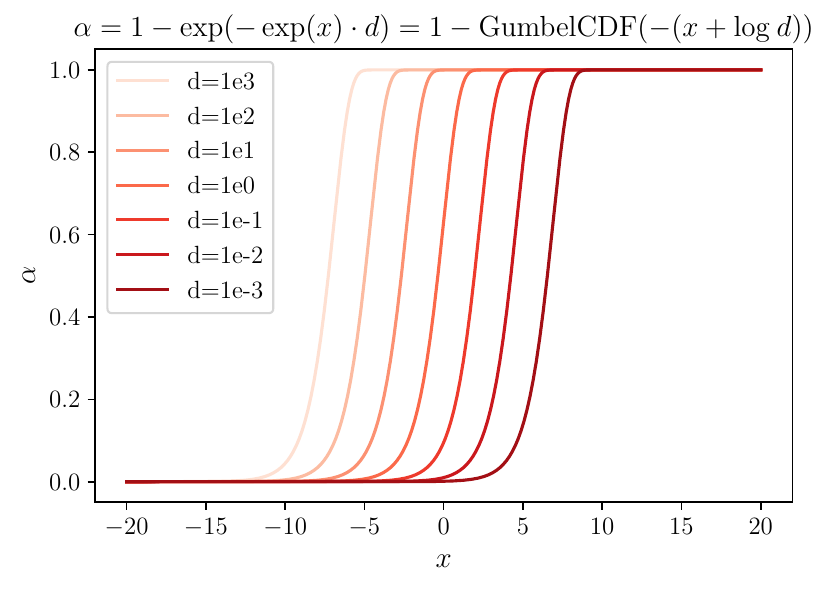}\\
\end{tabular}
\vspace{-1.5em}
\caption{
    $\alpha$ as a function of the raw input $x$ and $d$. We focus on $\alpha$ as a function of $x$, with different activation functions $\sigma(x)$, for a set of fixed values of interval lengths $d$. The $\mathtt{exp}$ activation (right plot) leads to a smooth, sigmoid-like transition from low to high $\alpha$ values regardless of the interval length $d$.
    The function $\exp(-\exp(-x))$ is the CDF of Gumbel distribution, and it is a numerically stable recipe that we recommend over $\mathtt{trunc\_exp}$ because log density and log distance naturally cancel out each other before exponentiation.
}
\label{fig:act}
\vspace{-1em}
\end{figure*}

In our testing of these models we identify two main failure modes. When the scene is scaled down (short ray interval $d$), some models struggle to produce large enough $\sigma$ values for solid geometry. When the scene is scaled up (long interval length), the $\sigma$ at initialization is often too large, resulting in cloudiness that traps the optimization at bad local optima. We therefore propose 1) parameterizing
both distance and volume densities in log space for easier multiplicative scaling; 2) a closed-form formula for $\sigma$ value initialization that guarantees high ray transmittance and scene transparency. We show that these two ingredients can robustly handle various scene sizes. We are not the first to use $\mathtt{exp}$ activation for volumetric densities.
It was described in Instant-NGP~\cite{instantngp} and adopted by many subsequent works.
However, existing informal discussions for why $\mathtt{exp}$ should be used are somewhat  unsatisfactory. We believe that a clearer understanding of its needs would be of benefit.
In addition, since distance and densities compensate for each other, moving both to log space
i.e. the GumbelCDF form naturally provides numerical stability without the need for truncated $\mathtt{exp}$~\cite{torchngp}.

\myparagraph{Our contributions:}
\begin{itemize}
\item We discuss the notion of alpha invariance, and clarify the relationship of inverse multiplicative scaling between volumetric densities $\sigma$ and scene size. It is a basic property that applies %
to volume rendering and radiance fields.
\item We survey and ablate popular NeRF architectures on their modeling decisions related to alpha invariance.
\item We provide a robust, general recipe that enables NeRFs to achieve consistent view synthesis quality across different scene scalings.
\end{itemize}

\section{Background}
\label{sec:background}

\paragraph{Volume rendering.}  In a space permeated by fog, it is unlikely for a faraway photon to arrive at the sensor. On a ray $\rvz(t) = \rvo + t \rvd$, the probability a photon comes from beyond $t$ i.e. $(t, \infty)$ is modelled by a decaying transmittance function $T(t) = \exp(-\int_0^t \sigma(s) \, ds) $, which vanishes with larger $t$.
The volume density $\sigma$ is some local light blocking factor, and the cumulative distribution function (CDF) is $1 - T(t)$. The prob. density $w(t)$ of a photon starting at \emph{exactly} time $t$ is thus ${\partial_t} (1-T(t))$. With an infinite number of photons, by law of large numbers, the deposited color at the pixel is the sample mean which converges to the expectation
\begin{equation}
\label{eq:cont_volrend}
\hspace{-.15cm}\mathbb{E}[\rvc(t)] = \int_{0}^{\infty} w(t) \, \rvc(t) \, dt = \int_{0}^{\infty} \underbrace{\sigma(t) \, T(t)}_{= {\partial_t} (1-T(t))} \rvc(t) \, dt.
\end{equation}
A NeRF optimizes a mapping from spatial location $\rvz$ to color and volume density $f_{\theta}: (\rvz, *) \rightarrow (\rvc, \sigma)$, where $*$ denotes auxiliary inputs such as viewing direction~\cite{nerf}, time~\cite{dnerf}, scene specific embeddings~\cite{nerfw}, etc.
The integral of Eq.~(\ref{eq:cont_volrend}) is approximated by assuming constant density $\sigma$ in a small interval of length $\Delta t$. CDF increment provides the weighting $w(\Delta t)|_t = T(t) - T(t + \Delta t) = T(t) - T(t)e^{-\sigma \Delta t} = e^{-\int_0^t \sigma(s) \, ds} \cdot (1 - e^{-\sigma \Delta t}) $. For segments each with length $d_i$ and density $\sigma_i$, $w_i = e^{-\sum_{j<i} \sigma_j d_j} \cdot (1 - e^{-\sigma_i d_i})$ is written as alpha compositing, %
\bea
\label{eq:alpha}
\alpha_i = 1 - e^{- \sigma_i  d_i} %
~\text{ with }~w_i = \prod_{j < i} (1 - \alpha_j) \cdot \alpha_i.
\eea
See~\figref{fig:volrend} for a tree-branching analogy.
The final color is the weighted sum $\sum_i w_i \, \rvc_i$.

\paragraph{Current practices on $\sigma$ parameterization.} A scalar $x$ predicted by the underlying neural field is converted to volume density $\sigma(x)$ by a non-linear activation function. The choice of this activation is one of the differences between NeRF models, and the focus of our analysis. It determines how $\alpha$ in~\equref{eq:alpha}, which is a function of both $d$ and $\sigma(x)$, depends on $x$.

MLP-NeRF~\cite{nerf} parameterizes the volume density $\sigma(x)$
with a $\mathtt{ReLU}$ activation. Mip-NeRF~\cite{mipnerf}
advocates for the use of $\mathtt{softplus}$ activation, out of concern that $\mathtt{ReLU}$ might get stuck since there is no gradient if inputs are negative. DVGO~\cite{dvgo} fixes the local \emph{interval} size. TensoRF~\cite{tensorf} scales the local interval size by a constant. We refer the readers to GitHub issues where these are discussed
\footnote{\url{https://github.com/sunset1995/DirectVoxGO/issues/7}}
\footnote{\url{https://github.com/apchenstu/TensoRF/issues/14}}
.
Plenoxels~\cite{plenoxels} uses $\mathtt{ReLU}$ with a very large learning rate on $\sigma(x)$ at the beginning
of optimization before decaying it as convergence improves\footnote{\url{https://github.com/sxyu/svox2/issues/111}}.
Instant-NGP uses $\mathtt{exp}$ activation. The motivation
is not explained, besides a brief comment by the author on GitHub\footnote{\url{https://github.com/NVlabs/instant-ngp/discussions/577}}. Some of these decisions have been adopted by more recent works. For example, HexPlane~\cite{hexplane} and LocalRF~\cite{localrf} follow TensoRF's interval scaling strategy, while works building on top of Instant-NGP tend to use the truncated $\mathtt{exp}$ activation for numerical stability.

\captionsetup[algorithm]{font=normal}
\begin{algorithm}[t]
\caption{Python pseudocode for our 1) GumbelCDF density activation and 2) high transmittance initialization. It is \emph{numerically stable} 
since log densities, log distances, and high transmittance offset naturally cancel out one another when scene size $L$ or the ray sampling strategy changes. 
}
\label{alg:density_shift_init}
\begin{lstlisting}
class DensityField():
    def __init__(self, L, tau=1.0, T=0.99):
        # L could be far - near of a typical light ray
        # tau: expected std of the neural field output
        self.offset = \
            log(log(1/T)) - log(L) - (tau**2)/2
        self.encoder = YourDensityEncoder()

    def compute_volrend_ws(self, xyz_locs, deltas):
        # xyz_locs, deltas: shape [N_rays, N_samp]
        log_densities = self.encoder(xyz_locs)
        density_delta = exp(
            log_densities 
            + log(deltas)   # GumbelCDF 
            + self.offset   # high transmittance
        )   
        # log(deltas) + offset invariant w.r.t. L
        trans = append_zeros_in_front(
            cumsum(density_delta[..., :-1], dim=-1), 
            dim=-1
        )
        trans = exp(-trans)
        alphas = 1. - exp(-density_delta)
        weights = alphas * trans
        return weights
\end{lstlisting}
\end{algorithm}

\section{Alpha Invariance}\label{sec:alphainv}

As reviewed in~\secref{sec:background}, $w(t) = \sigma(t) T(t)$ forms a probability density function. However, due to the scale-ambiguity of a 3D scene, the size of the support for $w(t)$ is arbitrary. For example in the original NeRF blender synthetic dataset,
a light ray travels over $[2.0, 6.0]$, corresponding to ray length (scene scale) $L=4.0$. If we were to expand the support by $2 \times$ to $L=8.0$, then by change of variables, it is \emph{sufficient} for the probability density $w(t)$ and thus the volume density $\sigma(t)$ to scale down by $\frac{1}{2}$ to integrate to 1 and render the same color.
Of course for a given 3D scene, no matter how the distance unit changes, the cumulative opacity of a fixed piece of ray segment is constant. In other words, a desired property of a model is that values of $\alpha$ in~\equref{eq:alpha} should be invariant with respect to (arbitrary) scene scaling. We refer to this desired property as \textbf{alpha invariance}.

The magnitude of volume density is also affected by the sampling resolution on
each ray. Most NeRF variants use some form of importance sampling
to iteratively focus the computation on the fine, detailed structures that would likely be missed by the initial uniform samples. Consider an extremely coarse sampling with only 2 sampled points: each
interval is unusually large, and the density needed to achieve a high alpha would be small.
On the other hand, with fine-grained importance sampling, large densities are needed to create a sharp, solid geometry within the small interval.
We can calculate the volume density required to achieve a certain alpha value by $-\frac{1}{d}\log(1 - \alpha)$. Some example values are shown in~\tabref{tab:anticipated_sigma_values}, where we set the overall ray length again to $L=4.0$
and vary $L$, the sampling resolution, and the alpha threshold.

\begin{table}[t]
\centering
\setlength{\tabcolsep}{4.7pt}
\renewcommand{\arraystretch}{1.06}
\begin{tabular}{lcccc}
\specialrule{.15em}{.05em}{.05em}
& $d = \frac{L}{64}$ & $d = \frac{L \times 2}{64}$ & $d = \frac{L}{128}$ & $d = \frac{L}{64 \times 128}$\\
\midrule
$\alpha = 0.5$ & 11.1 & 5.5 & 22.2  & 1419.6 \\ 
$\alpha = 0.99$ & 73.7 & 36.8 & 147.4 & 9431.4\\ 
$\alpha = 0.999$ &  110.5 & 55.3  & 221.0  & 14147.1      \\
\specialrule{.15em}{.05em}{.05em}
\end{tabular}
\vspace{-0.15cm}
\caption{
Some example $\sigma$ values given desired alpha level and interval size,
using $\sigma = -\frac{1}{d}\log(1 - \alpha)$. Length of ray $L$ is set to $4.0$, which is blender synthetic dataset default, and we vary the number of samples per ray. 
Doubling $L$ halves the $\sigma$, while higher sampling resolution increases $\sigma$. In the extreme 
case where all the $128$ importance samples fall within one of the initial $64$ uniformly sampled bins, 
the magnitude of $\sigma$ can get very large.
}
\vspace{-0.2cm}
\label{tab:anticipated_sigma_values}
\end{table}

The three activation functions mentioned in~\secref{sec:background} as used in practice to compute $\sigma(x)$ lead to the following form for $\alpha$ as a function of $x$ and $d$:
\begin{subequations}
\begin{align}
\mathtt{ReLU}&: \alpha = 1 - \exp(-\mathtt{ReLU}(x) \cdot d) \label{eq:relu}\\
\mathtt{softplus}&: \alpha = 1 - \mathtt{sigmoid}(-x)^d \label{eq:softplus}\\
\mathtt{exp} &: \alpha = 1 - \exp(-\exp(x) \cdot d)\label{eq:exp}
\end{align}
\end{subequations}
where we use the identity $\exp(-\mathtt{softplus}(x)) = \mathtt{sigmoid}(-x)$.  %
To more clearly compare the effect of $\mathtt{exp}$ against $\mathtt{ReLU}$ and $\mathtt{softplus}$, in~\figref{fig:act} we visualize the behavior of $\alpha(x,d)$ with these activations, for different segment length $d$ from $10^{-3}$ to $10^{3}$.
When $d$ is small, both $\mathtt{ReLU}$ and $\mathtt{softplus}$ struggle to produce large $\alpha$ values; the values of $x$ that the networks must predict become extreme. In contrast, with the $\mathtt{exp}$ activation $\sigma=\exp(x)$ a constant offset on $x$ results in a direct scaling on $\sigma$.

We also note that the function $e^{-e^{-x}}$ is the CDF of Gumbel distribution,
\begin{align}
\label{eq:gumbel}
	\alpha &= 1 - \exp(-\exp(x) \cdot d) \nonumber \\
	&= 1 -  \exp(-\exp(x + \log d) ) \nonumber \\
	&= 1 - \mathtt{GumbelCDF}(-(x + \log d)).
\end{align}
This connection is related to the fact that the volume rendering equation is a restatement of the Exponential distribution. The PDF and CDF functions in~\equref{eq:cont_volrend} are the PDF and CDF of the Exponential distribution assuming a constant ``rate''.
This can aid in numerical stability, a commonly cited issue with using $\exp$ to parametrize $\sigma$. Large density $\sigma$ is needed primarily to create sharp opacity change in a small distance interval.
If we move the distance multiplication into the $\mathtt{exp}$ as a $\log d$ addition, then there should not be an overflow issue. Note that the derivative of $\exp(-\exp(x))$ is not symmetrical about y-axis even though its shape resembles $\mathtt{sigmoid}$.

While $\mathtt{exp}$ activation has no problem producing large densities when scene scale $L$ is small, a different challenge occurs when $L$ is large. Given the same sampling strategy, larger $L$ and therefore larger interval $d$, with the same initial volume density will produce larger alpha values. This manifests as an opaque initial scene. Optimization frequently fails since the images can be explained away by dense, cloudy floaters in front of each camera. It is important
to guarantee that upon initialization the scene is transparent,~\ie each ray has high transmittance.
\begin{figure}[t]
\centering
\includegraphics[width=\linewidth]{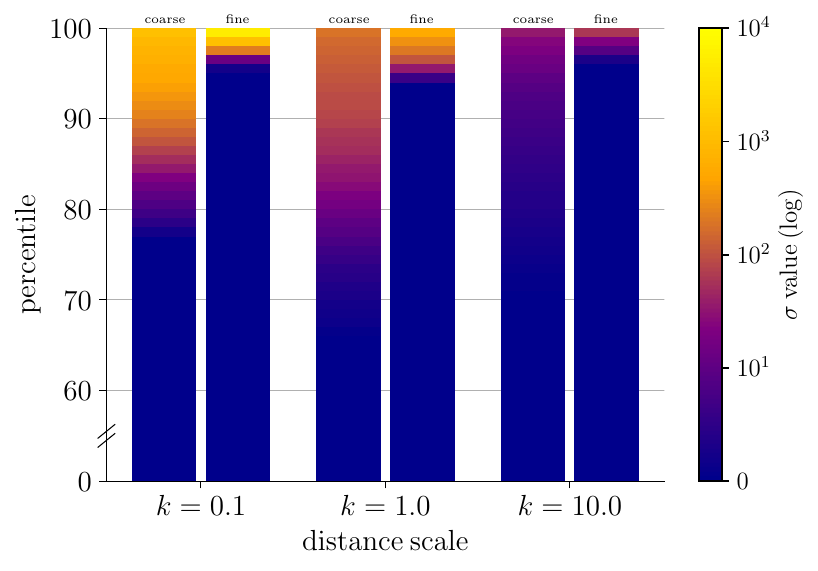}
\vspace{-0.6cm}
\caption{
    Distribution of volume density $\sigma$ in the lego bulldozer scene, queried via a uniformly sampled dense grid of points from both the coarse (left columns) and fine (right columns) MLP networks of vanilla-NeRF for different $k$.
    Color represents the average $\sigma$ value in each percentile of the sorted $\sigma$ distribution.
    The fine MLPs produce larger $\sigma$ than the coarse MLPs, and as $k$ increases, the magnitude of $\sigma$ decreases for both networks.
}
\label{fig:lego_cdf_boxplot}
\vspace{-0.23cm}
\end{figure}

Since ray transmittance is related to volume density by $T(L) = \exp(-\int_0^L \sigma(s) \, ds) $, we may want the network predicting the density to be initialized so that the ``average volume transmittance'' in the scene is $T'$, a value like $0.99$. In practice the initial volume density value along a ray is not a constant, but a random variable produced by the underlying field function such as voxels or an MLP. We make the simplifying assumption that the sampled values are i.i.d. (admittedly imperfect, since neighboring values tend to be correlated by feature or voxel sharing). The integral $-\int_0^L \sigma(s) \, ds$ gives the Monte Carlo estimate that converges to $-L \cdot \mathbb{E} [\sigma] $. Our recommendation can then be written as
\begin{equation}
    \exp(-L \cdot \mathbb{E}[\sigma]) = T'.\label{eq:recommend1}
\end{equation}
In case of $\sigma(x) = \mathtt{exp}(x)$, assuming the pre-activation field output $x$ is drawn from $\mathcal{N}(\mu, \tau^2)$, $\sigma$ follows a log-normal distribution with mean $\mathbb{E} [\sigma] = \exp(\mu + \frac{\tau^2}{2}) $.
Plugging this into~\equref{eq:recommend1}, taking log, and rearranging, we get
\begin{equation}
    \exp(\mu + \frac{\tau^2}{2}) = \frac{1}{L} \log \frac{1}{T'},\label{eq:recommend2}
\end{equation}
and solving for $\mu$ we obtain
\begin{equation}
\label{eq:init}
\mu =  \log \log \frac{1}{T'} - \log L - \frac{\tau^2}{2}.
\end{equation}
This is the desired target for initializing the density-predicting network.
Setting the mean $\mu$ of the pre-activation field output to this value can be done with an additive offset. This strategy is discretization agnostic, and a factor multiplied onto $L$ can be undone by a logarithmic shift. The merged expression for local alpha as a function of field output $x$ and interval length $d$ is
\bea
\alpha =&
\hspace{-2.65cm}1 -  \exp(-\exp(x + \log d + \mu) ) \nonumber \\
 =& \hspace{-.15cm}1 -  \exp \left( -\exp(x + \log \frac{d}{L} + \log \log \frac{1}{T'} - \frac{\tau^2}{2}) \right),
\eea
where $\log{\frac{d}{L}}$ is invariant w.r.t. scene scaling. See~\algref{alg:density_shift_init} for Python pseudocode.

For a real-life scene where ray length varies, a possible choice is to set $L$ to the longest ray distance. One interpretation of $\log{\frac{d}{L}}$ is that we are working with distance ratios and are thus effectively hardcoding the overall scene size to $1.0$. But it's only possible with the extra offset terms, without which, assuming $\tau=1.0,~ T'=0.99$, the scene size $L$ needs to be set to $0.006$.

Analogous recipes can be made for $\mathtt{ReLU}$ and $\mathtt{softplus}$, although it's harder to write down closed-form expressions~\cite{wiki:rectified_gaussian}.
We could attempt it using Monte Carlo estimates of their mean. Recall our goal is to achieve $\E[\sigma] \cdot L = \log(\frac{1}{T'})$. Assuming the pre-activation neural field output is drawn from $\mathcal{N}(0, 1)$, the expectations of the post-activation random variables can be approximated by sample mean. When $T'=0.99$, the RHS is $0.001$. We want to \emph{avoid} directly scaling down density or L to meet this target, since that would make it even harder for $\mathtt{relu}$ and $\mathtt{softplus}$ to achieve large density needed at solid regions. Instead the high transmittance should be achieved by shifting the activation function profile to the right, i.e. adding negative offset to the input. The correct offset can be tuned numerically, and only needs to be done once.

\begin{figure*}[t]
\centering
\includegraphics[width=\linewidth]{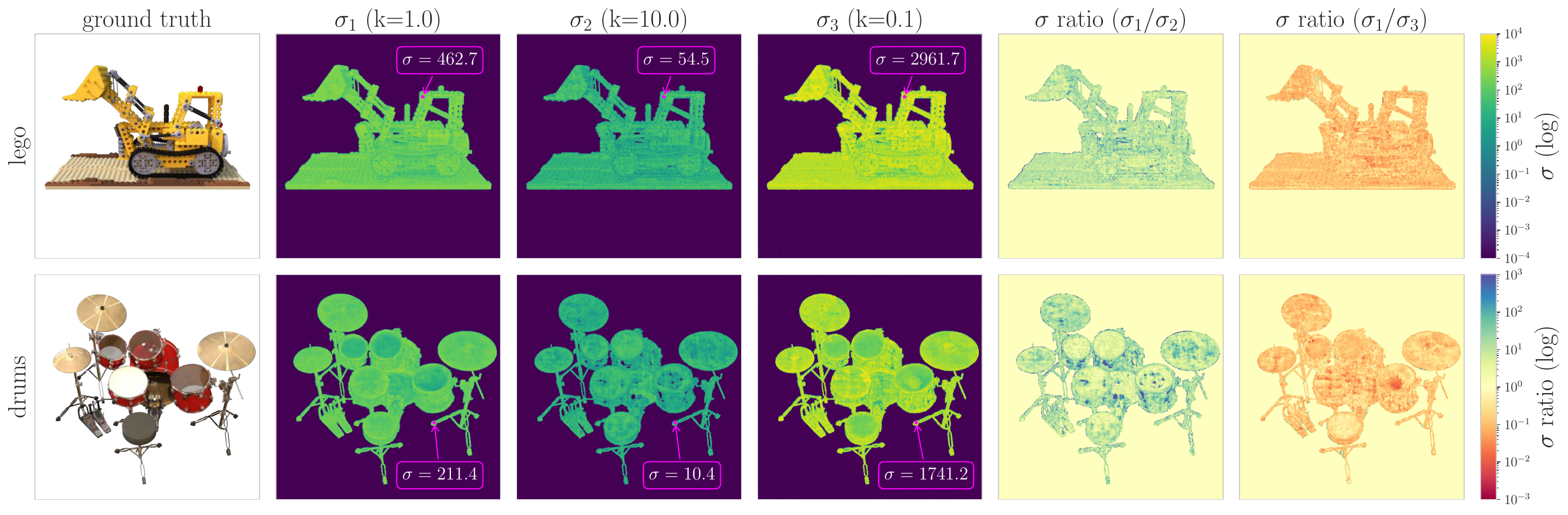}\\
\vspace{-0.1cm}
\includegraphics[width=\linewidth]{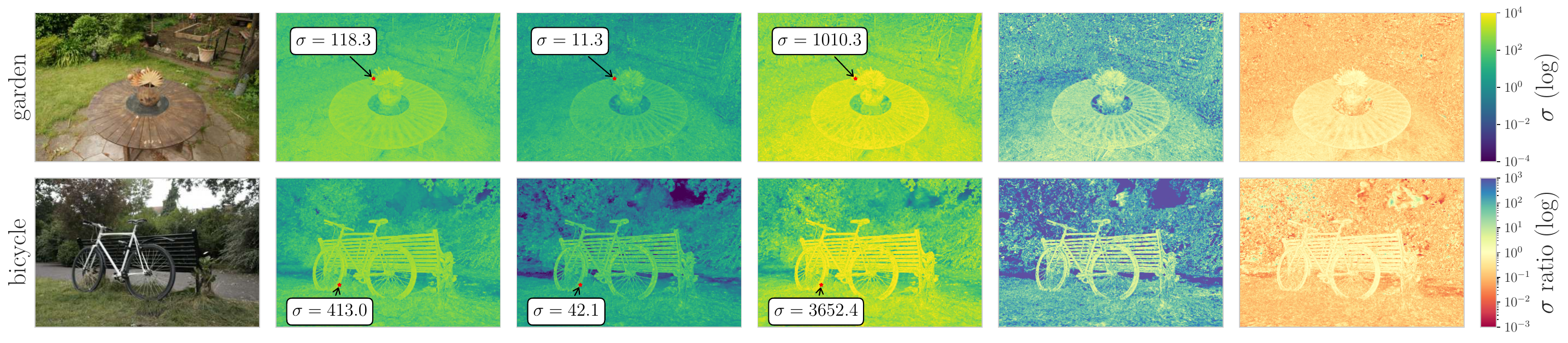}
\vspace*{-0.75cm}
\caption{
Values of volume density $\sigma$ on the object surface, with models trained under different scene scalings $k$. On each ray, surface point is defined as the 50th percentile location of volume rendering CDF. We annotate a few prominent points, and also produce two $\sigma$ division images that show the overall ratio of the numerical range of $\sigma$ at different scaling factors.
The blender scenes are trained with vanilla NeRF, and the Mip-NeRF 360 scenes are trained with Nerfacto. The ratio of $\sigma$ is empirically close to $1/k$.
}
\label{fig:all_50_percentile_sigma_viz}
\vspace{-0.5cm}
\end{figure*}

\section{Experiments}

Scene size $L$ is fundamentally an arbitrary decision for NeRF optimization.
A robust algorithm should be able to achieve consistent view synthesis quality regardless of the value of $L$.
We analyze different types of NeRF systems including pure MLPs to Voxels to Hashgrid-MLP hybrids by training them with a scaling factor $k$ applied on the default scene size $L$, and leave the rest of the optimization hyperparameters such as sampling strategies unchanged.
In our experiments, unless otherwise stated, we set the desired transmittance $T' = 99\%$ for the longest ray in each scene, and set $k$ to range from $0.1$ to up to $25$, report the PSNR metrics, and provide qualitative visualizations.
We find that these methods are unable to maintain high rendering quality especially when $k$ is more extreme, whereas our recipe achieves a consistently high quality across all $k$.

\subsection{How volume density $\sigma$ changes in practice}
\begin{mdframed}[style=MyFrame,align=center]
\begin{findings}
\vspace{-0.15cm}
Empirically $\sigma$ changes by a factor close to $\frac{1}{k}$ when scene size is changed by $k$. It is most noticeable on object surfaces.
\end{findings}
\end{mdframed}
\vspace{-0.25cm}
Inverse scaling between volume density and distance is sufficient to guarantee
identical renderings for a radiance field. However, this might not be necessary in practice if the end goal is to achieve high PSNR values. A scene is typically dominated by either empty space or solid regions; semi-transparent structures occur much less frequently, and volumetric densities tend to take on extreme values. As such, a change in the scale of the interval length might not substantially alter the local alpha values. To get a better understanding of their exact empirical behaviors, we train vanilla-NeRF on the lego and drums scenes from the blender dataset, and Nerfacto~\cite{nerfstudio} on the garden and bicycle scenes from the Mip-NeRF 360 dataset, with $k=[0.1,1.0,10.0]$. Here we use our proposed~\algref{alg:density_shift_init}.

\paragraph{Volume statistics.}
For the coarse and fine MLP networks in vanilla-NeRF,
we query a dense grid of $200^3$ uniformly sampled points in the scene box,
and summarize the sorted $\sigma$ distribution with bar plots depicting the average
$\sigma$ value within each percentile in
~\figref{fig:lego_cdf_boxplot}.
As most of the scene is empty ($>60\%$), we focus on the distribution of values in the upper percentiles and observe that with increasing $k$ values, the numerical range of $\sigma$ does decrease as we hypothesized in both the coarse and fine MLP networks. In addition, we observe that the fine MLPs produce a very small amount of significantly larger $\sigma$ values than the coarse MLPs due to sharper geometry being captured by the importance sampler over the course of optimization.

\paragraph{Surface statistics.}
Since large volume density occurs near the object surface, we visualize the changing surface densities in~\figref{fig:all_50_percentile_sigma_viz}.
At a given camera pose, we shoot rays through each pixel to obtain the density histograms $\{w_i\}$, and query the spatial location at the $50$-th percentile of the probability CDF of each histogram for its $\sigma$ value. We also display heatmaps of density ratio by dividing the $\sigma$ images coordinate-wise.
Values below $\epsilon=10^{-4}$ are rounded up to $\epsilon$ for numerical stability.
Based on these visualizations, we observe that inverse-scaling between density and distance does hold to a significant degree, although the factor in practice deviates from $1/k$.

\subsection{Vanilla NeRF with MLPs}

\begin{table*}[t]
\centering
\setlength{\tabcolsep}{4pt}
\small
\begin{tabular}{lcccccccc}
\specialrule{.15em}{.05em}{.05em}
 & \textbf{chair} & \textbf{drums} & \textbf{ficus} & \textbf{hotdog} & \textbf{lego} & \textbf{materials} & \textbf{mic} & \textbf{ship} \\
\hline
$k = 0.1$ & 30.98 / 31.19 & 24.25 / 24.37 & 28.72 / 29.26 & \textcolor{red}{32.22} / \textcolor{blue}{34.84} & 30.90 / 30.64 & 28.27 / 28.45 & \textcolor{red}{28.31} / \textcolor{blue}{30.96} & 26.70 / 27.56 \\
$k = 0.4$ & 31.21 / 31.29 & 23.85 / 24.54 & 28.74 / 29.01 & 32.95 / 33.51 & \textcolor{red}{9.45} / \textcolor{blue}{30.71} & 28.02 / 28.59 & 31.41 / 30.94 & \textcolor{red}{10.45} / \textcolor{blue}{27.35} \\
$k = 1.0$ & 31.26 / 31.26 & \textcolor{red}{12.04} / \textcolor{blue}{24.58} & 28.76 / 28.85 & 33.59 / 33.71 & 30.71 / 31.36 & \textcolor{red}{13.89} / \textcolor{blue}{28.57} & 31.35 / 31.05 & \textcolor{red}{25.65} / \textcolor{blue}{27.21} \\
$k = 2.5$ & 31.22 / 31.32 & 24.04 / 23.99 & \textcolor{red}{26.86} / \textcolor{blue}{28.60} & \textcolor{red}{10.94} / \textcolor{blue}{33.91} & 30.74 / 31.11 & 27.76 / 28.85 & 30.97 / 31.35 & \textcolor{red}{5.88} / \textcolor{blue}{27.15} \\
$k = 10.0$ & \textcolor{red}{14.04} / \textcolor{blue}{31.36} & 23.84 / 24.43 & 28.97 / 28.92 & \textcolor{red}{10.39} / \textcolor{blue}{33.23} & 30.55 / 30.84 & 27.97 / 28.25 & 31.04 / 31.44 & \textcolor{red}{5.88} / \textcolor{blue}{26.99} \\
\specialrule{.15em}{.05em}{.05em}
\end{tabular}
\vspace{-0.2cm}
\caption{
PSNR $\uparrow$ for NeRF-Pytorch [\textcolor{red}{baseline} / \textcolor{blue}{ours}] at different scene scaling $k$ on Blender synthetic dataset. Vanilla NeRF with $\mathtt{ReLU}$ activation is surprisingly capable of producing large $\sigma$ values
but the optimization does converge to poor local minima randomly. Using %
our recommended recipe ensures consistent convergence across all $k$.
Note that the NeRF-Pytorch codebase is unable to exactly match the original NeRF~\cite{nerf} performance at 200k iterations. See appendix for details on reproducing the random failures present here.
}
 \vspace{-0.2cm}
\label{tab:vanilla-nerf-blender}
\end{table*}

\begin{mdframed}[style=MyFrame,align=center]
\begin{findings}
\vspace{-0.15cm}
MLPs are surprisingly robust with just $\mathtt{ReLU}$ activation but they can be improved.
\end{findings}
\end{mdframed}
\vspace{-0.25cm}
The canonical NeRF architecture uses an 8-layer MLP with a $\mathtt{ReLU}$ activation
to produce $\sigma$, and hence must output values in the range of hundreds
near object surfaces as shown in ~\figref{fig:all_50_percentile_sigma_viz}. This setup is somewhat against the conventional wisdom~\cite{bishop_book,dnn_normalization} of fixing the neural network output magnitude to unit variance for stable training.
We test our recipe on NeRF-PyTorch~\cite{lin2020nerfpytorch}, train for 200k steps and present the results in~\tabref{tab:vanilla-nerf-blender}.
We make three observations.
First, NeRF-PyTorch uses PyTorch's default linear layer initialization which does not correct for the variance gain of $\mathtt{ReLU}$, and produces raw outputs with mean $0$ and variance $0$. There are random failure modes across all $k$.
To address this, we initialize the layers with Kaiming uniform initialization~\cite{he_init} with $\sqrt{2}$ gain. Interestingly, we still observe random failure modes, even at $k=1$.
Finally, we tried parameterizing density with our recipe (without applying the aforementioned Kaiming uniform initialization such that $\tau=0.0$ initially, making the task more difficult), and observed consistent performance across all $k$ without any random failures. %

\subsection{Voxel Variants: DVGO, Plenoxels and TensoRF}\label{sec:voxel_variants}
\begin{figure}[t]
\centering
\includegraphics[width=0.95\linewidth]{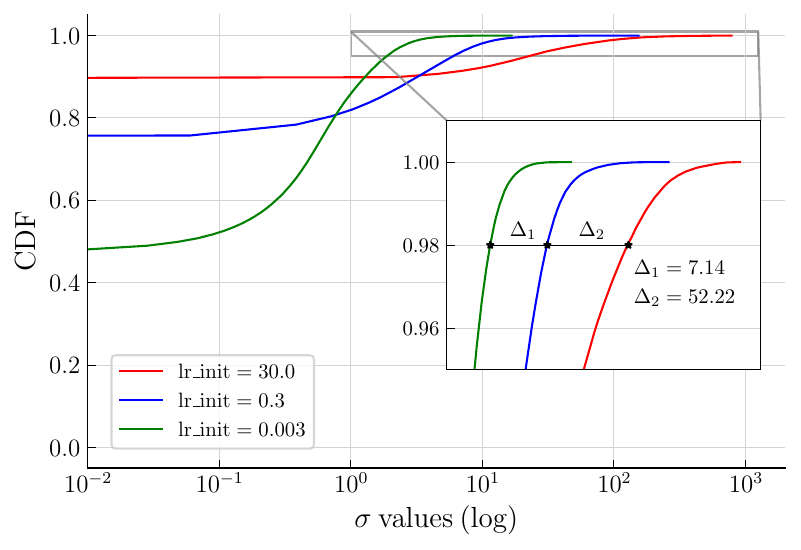}
\vspace{-0.25cm}
\caption{
     CDF of the $\sigma$ distributions produced by Plenoxels on the T-Rex scene from the LLFF dataset with different learning rate schedules on the $\sigma$ voxels. The default high learning rate schedule \textcolor{red}{[red]} is needed to produce large $\sigma$ values and high PSNR. Lower learning rates lead to smaller $\sigma$ and ultimately worse performance.
}
\label{fig:plenoxels_cdf}
\vspace{-0.25cm}
\end{figure}

\begin{figure}[t]
\centering
\includegraphics[width=0.9\linewidth]{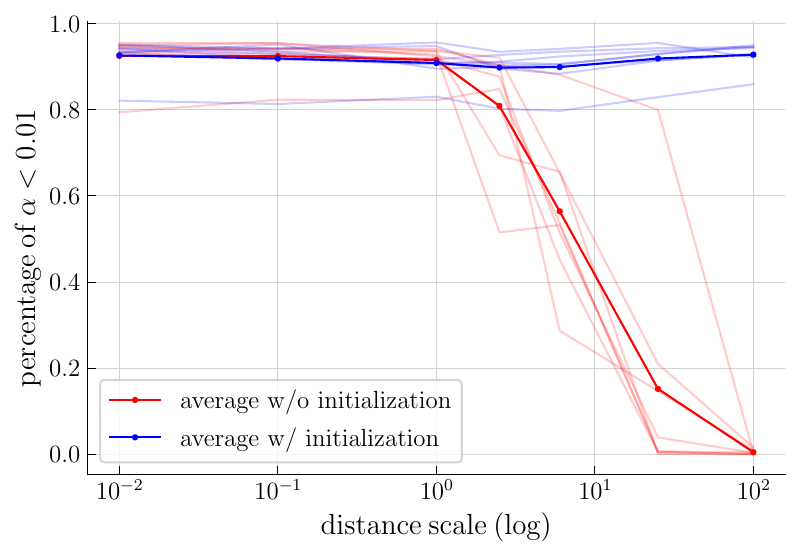}
\vspace{-0.3cm}
\caption{
    Querying a dense uniform grid of samples from TensoRF trained with $\exp$ activation. We consider a location to be empty if the local alpha is below $0.01$, and apply distance scaling $k$ ranging from $10^{-2}$ to $10^{2}$.
    Each faint line corresponds to a scene from the blender dataset; the solid line is the dataset average. With our high transmittance initialization, we prevent over-densification even when $k$ is large.
}
\vspace{-0.25cm}
\label{fig:tensorf_init_alpha_comp}
\end{figure}

\begin{mdframed}[style=MyFrame,align=center]
\begin{findings}
\vspace{-0.15cm}
Voxel variants cannot converge with $\mathtt{ReLU}$ or $\mathtt{softplus}$ activations alone. Hardcoded heuristics have been used for stable training.
\end{findings}
\end{mdframed}
\vspace{-0.25cm}
Unlike MLPs, direct optimization of voxel NeRFs with $\mathtt{ReLU}$ or $\mathtt{softplus}$ activations does not converge. We observe the same pattern across all three voxel model variants, and it shows the benefit of having an MLP as a global inductive bias. DVGO uses $\mathtt{softplus}$ activation, and to address the issue of divergence, it fixes a canonical scene size with the voxel size ratio such that each \emph{interval} has length $0.5$ or $1.0$ depending on whether it is in the coarse or fine reconstruction stage. It then calculates a density offset such that every \emph{local} cell's alpha value is small at initialization. We instead recommend initializing the density network based on overall ray transmittance.
TensoRF applies a constant interval scaling of $25$ with $-10$ offset on $\mathtt{softplus}$ input. We verify the robustness of our overall recipe in comparison with TensoRF's and provide numerical metrics in~\tabref{tab:tensorf_compare_blender}.
We also ablate our initialization offset on TensoRF.
Applying $\mathtt{\exp}$ alone is not enough to make TensoRF robust across scene scalings; as $k$ increases, we show that the distribution of alpha values everywhere in the scene tends towards $1$. Applying our high transmittance offset enables TensoRF to converge smoothly, regardless of $k$.
See~\figref{fig:tensorf_init_alpha_comp}.
Plenoxels uses a total variation (TV) regularizer, and $\mathtt{ReLU}$ activation with an unusually large initial learning rate of $30.0$, in addition to a reverse cosine decay on the $\sigma$ coefficients. Disabling it, \ie reducing the learning rate to a lower value such as $0.3$ or $0.003$, produces smaller $\sigma$, and is directly correlated with worse rendering quality. See~\figref{fig:plenoxels_cdf} and~\tabref{tab:plenoxels_psnr}.

\begin{table*}[t]
\centering
\setlength{\tabcolsep}{3.9pt}
\small
\begin{tabular}{lcccccccc}
\specialrule{.15em}{.05em}{.05em}
 & \textbf{chair} & \textbf{drums} & \textbf{ficus} & \textbf{hotdog} & \textbf{lego} & \textbf{materials} & \textbf{mic} & \textbf{ship}\\ 
\hline
$k = 0.1$ & \textcolor{red}{fail} / \textcolor{blue}{35.41} & \textcolor{red}{fail} / \textcolor{blue}{26.01} & \textcolor{red}{fail} / \textcolor{blue}{33.81} & \textcolor{red}{fail} / \textcolor{blue}{37.03} & \textcolor{red}{fail} / \textcolor{blue}{36.47} & \textcolor{red}{fail} / \textcolor{blue}{29.99} & \textcolor{red}{fail} / \textcolor{blue}{34.52} & \textcolor{red}{fail} / \textcolor{blue}{30.51}  \\
$k = 0.4$ & 35.74 / 35.96 & 26.04 / 25.91 & \textcolor{red}{fail} / \textcolor{blue}{34.00} & 37.31 / 36.93 & 36.20 / 35.93 & 30.14 / 30.10 & \textcolor{red}{fail} / \textcolor{blue}{34.31} & 30.81 / 30.09     \\
$k = 1.0$ & 35.55 / 35.57 & 26.24 / 26.24 & 33.99 / 34.31 & 37.31 / 37.20 & 36.60 / 36.55 & 30.01 / 30.11 & 34.59 / 34.71 & 30.65 / 30.59   \\
$k = 2.5$ & 35.71 / 35.88 & 26.26 / 26.14 & 34.05 / 34.15 & 37.09 / 37.14 & 36.66 / 36.51 & 30.11 / 30.21 & 34.48 / 34.61 & 30.71 / 30.80     \\
$k = 10.0$ & 35.67 / 35.92 & 26.14 / 26.41 & 34.01 / 34.81 & 37.10 / 37.51 & 36.18 / 37.04 & 30.36 / 30.40 & 34.53 / 34.62 & 30.74 / 30.80    \\
\specialrule{.15em}{.05em}{.05em}
\end{tabular}
\vspace{-0.2cm}
\caption{
PSNR values of TensoRF [\textcolor{red}{baseline} / \textcolor{blue}{ours}] at different scene scaling $k$ on Blender synthetic dataset. 
TensoRF applies an input offset $-10$ before $\mathtt{softplus}$, followed by an interval multiplier of $25$. The hardcoded decisions are less effective at $k=0.1$. TensoRF is quite robust at $k=10$ since the $-10$ is a rather aggressive offset. See Fig~\ref{fig:act} for the effect of shifting the $\mathtt{softplus}$ activation. 
}
\label{tab:tensorf_compare_blender}
\vspace{-0.1cm}
\end{table*}

\subsection{MLP and Hashgrid Hybrids}

\begin{figure*}[t]
\centering
\includegraphics[width=\linewidth]{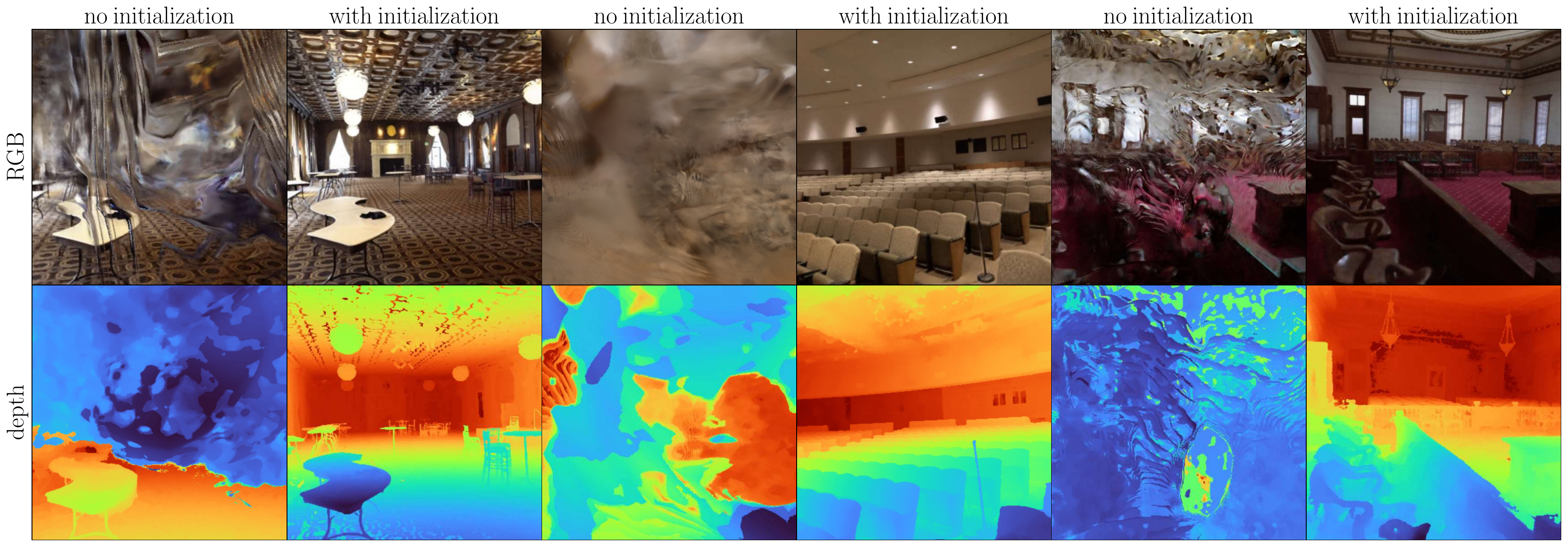}
\includegraphics[width=\linewidth]{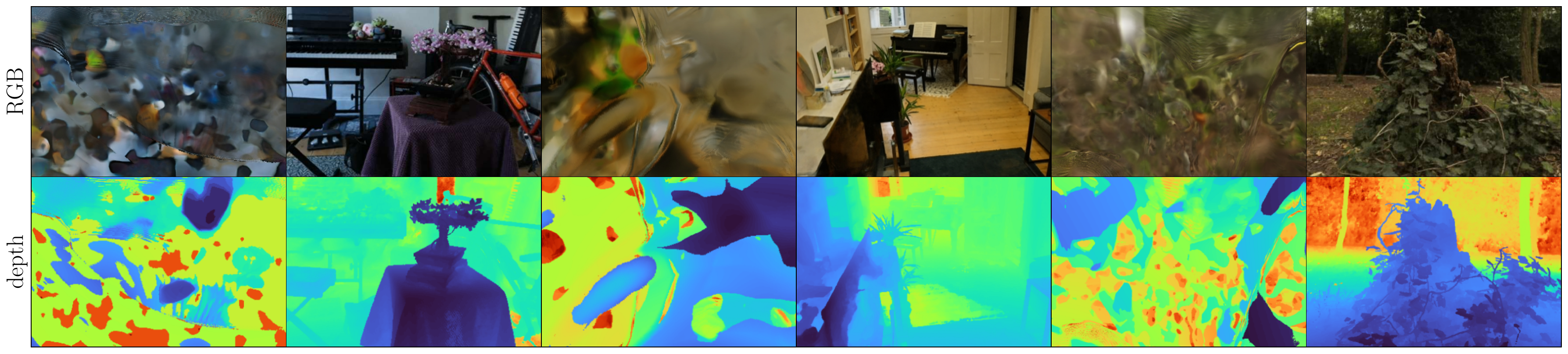}
\vspace{-0.7cm}
\caption{
    Image and depth maps produced by Nerfacto on the ballroom, auditorium, and courtroom scenes from Tanks and Temples at $k=25$, and the bonsai, room, and stump scenes from the Mip-NeRF 360 dataset at $k=10$. Not using high transmittance offset causes the optimization to get stuck with cloudy floaters.
}
\label{fig:exp_init_rgbd_nerfacto}
\vspace{-0.2cm}
\end{figure*}

\begin{mdframed}[style=MyFrame,align=center]
\begin{findings}
\vspace{-0.15cm}
The MLP$ + $Hashgrid hybrid with $\exp$ activation makes the model robust at small $k$. However, high transmittance offset is needed for large $k$.
\end{findings}
\end{mdframed}
\vspace{-0.25cm}
Instant-NGP~\cite{instantngp} and many follow-ups, including Nerfacto~\cite{nerfstudio}, already use $\mathtt{\exp}$ activation to parameterize $\sigma$. Here we study the Nerfacto architecture, since it subsumes many of Instant-NGP's components and supports additional features like Mip-NeRF 360's scene contraction~\cite{mipnerf360}.
While Nerfacto is robust at producing large $\sigma$ when $k$ is small, the optimization gets stuck in the ``cloudiness trap'' when $k$ is large, since the increased ray distances cause the
alpha values to increase, leading to an initially opaque scene.
See~\tabref{tab:nerfacto_360}
for experimental results on the Mip-NeRF 360
dataset.
See~\figref{fig:exp_init_rgbd_nerfacto} for RGB-image and depth-map comparisons.
Our high transmittance initialization strategy enables smooth optimization across $k$.

\begin{figure}[t]
\centering
\includegraphics[width=\linewidth]{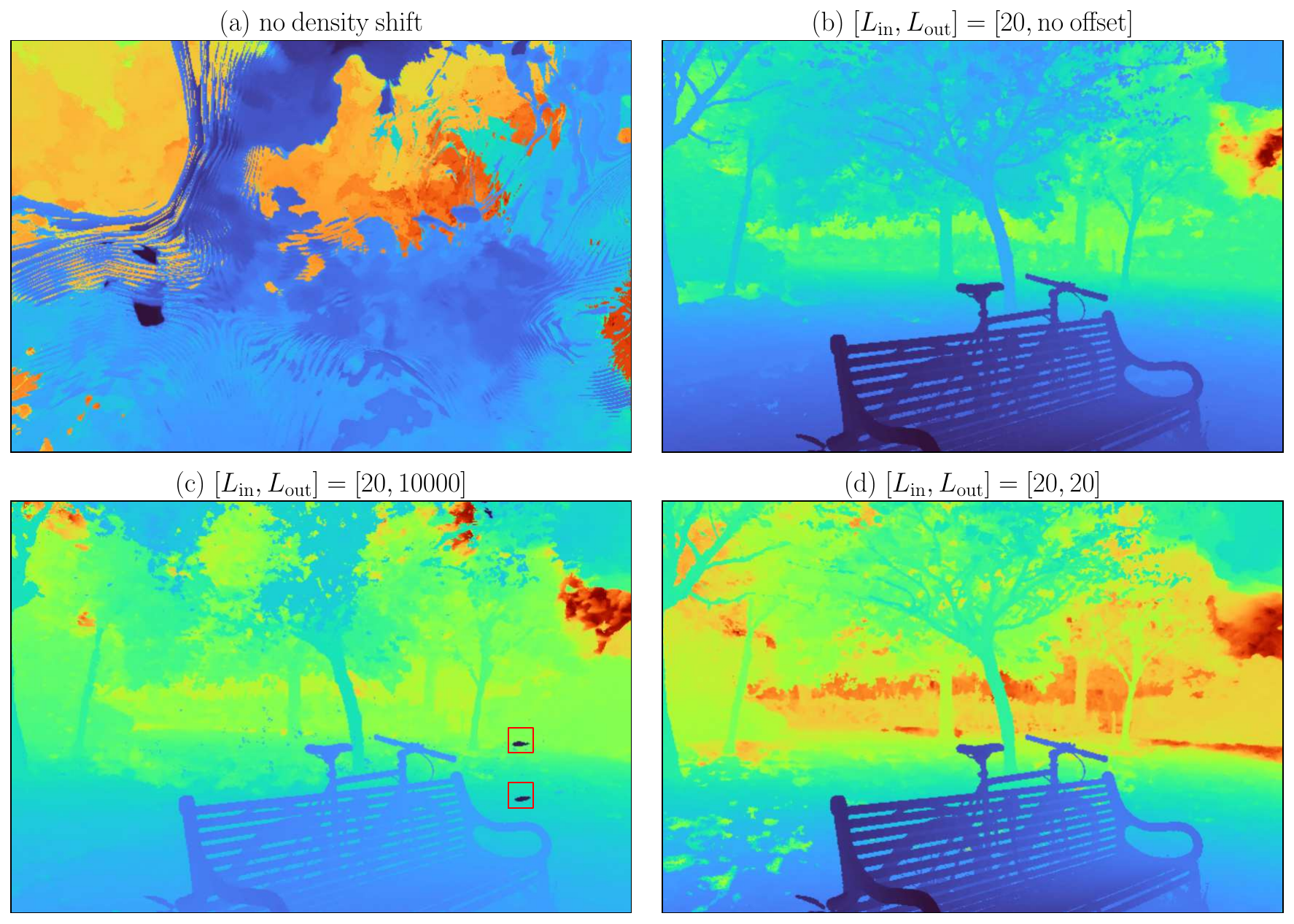}
\vspace{-0.6cm}
\caption{
    Depth maps from Nerfacto with various high transmittance offsets applied on a scene with contracted background. The dataset is ``Bicycle'' from the Mip-NeRF 360 with scaling $k=10$. (a) No transmittance offset on foreground or background leads to divergence. (b) Applying the offset only on the inner dome is fine for reconstructing the foreground object, but leaves a ``wall'' of opacity on the background. (c) Setting $L_\text{out}$ to be the metric ray distance produces misty floaters in near-camera regions. (d) $L_\text{out}= L_\text{in}$ performs well and is easy to implement.
}
\label{fig:scene_contraction}
\vspace{-0.50cm}
\end{figure}

\definecolor{tabfirst}{rgb}{1, 0.7, 0.7} %
\definecolor{tabsecond}{rgb}{1, 0.85, 0.7} %
\definecolor{tabthird}{rgb}{1, 1, 0.7} %

\begin{table*}[t]
\centering
\small
\begin{tabular}{lcccccccc}
\specialrule{.15em}{.05em}{.05em}
& \textbf{bicycle} & \textbf{bonsai} & \textbf{counter} & \textbf{garden} & \textbf{kitchen} & \textbf{room} & \textbf{stump} & \textbf{average} \\
\midrule
$[L_\mathrm{in}, L_\mathrm{out}] = [20, 10000]$   &  \textcolor{red}{19.81} &  27.34 &                      23.51 &  26.71 &                      27.46 & 28.60 & \textcolor{red}{19.57} & 24.71 \\
$[L_\mathrm{in}, L_\mathrm{out}] = [20, 40]$       &  22.27 & 28.41 &  25.19 & 26.23 &  28.23 & 28.72 &  23.74 & 26.11 \\
$[L_\mathrm{in}, L_\mathrm{out}] = [20, 20]$        & 22.11 &  28.94 & 25.66 & 26.00 & 28.14 &  29.03 & 24.02 & 26.27 \\
$[L_\mathrm{in}, L_\mathrm{out}] = [20, 0]$        & 22.25 &  28.46 & 25.03 &  26.11 & 27.77 &  28.44 & \textcolor{red}{18.42} & 25.21 \\
\specialrule{.15em}{.05em}{.05em}

\end{tabular}
\vspace{-0.2cm}
\caption{
PSNR $\uparrow$ for Nerfacto on the Mip-NeRF 360 dataset at $k=10$. $L_\text{in} = 2 \cdot k = 20$ (ignoring the small near-plane) and we explore different $L_\text{out}$ for transmittance offset in the outer dome. 
The metric ray distance $10000$ removes the background inductive bias as described in Sec~\ref{sec:scene_contract} and degrades performance. $L_\text{out} = L_\text{in}$ is the easiest to implement and has good performance. 
}
\label{tab:scene_contraction}
\vspace{-0.25cm}
\end{table*}

\begin{table}[t]
    \centering
    \small
    \setlength{\tabcolsep}{4.5pt}
    \begin{tabular}{lccc}
    \specialrule{.15em}{.05em}{.05em}
       & $\text{lr\_init}=30.0$ & $\text{lr\_init}=0.3$ & $\text{lr\_init}=0.003$ \\
    \hline
    blender & 31.66 & 29.51 & 23.51 \\
    LLFF~\cite{llff} & 24.51 & 23.22 & 21.09 \\
    \specialrule{.15em}{.05em}{.05em}
    \end{tabular}
    \vspace{-0.25cm}
    \caption{Measuring Plenoxels' performance with PSNR $\uparrow$ on the Blender and LLFF datasets with different learning rate schedules.}
    \label{tab:plenoxels_psnr}
    \vspace{-0.25cm}
\end{table}

\begin{table*}[t]
\centering
\setlength{\tabcolsep}{3.9pt}
\small
\begin{tabular}{lccccccc}
\specialrule{.15em}{.05em}{.05em}
 & \textbf{bicycle} & \textbf{bonsai} & \textbf{counter} & \textbf{garden} & \textbf{kitchen} & \textbf{room} & \textbf{stump} \\
 \midrule
$k = 0.1$ & 22.51 / 22.38 & 28.71 / 28.86 & 25.22 / 25.16 & 26.26 / 26.63 & 28.04 / 28.24 & 28.70 / 28.71 & 23.42 / 23.48 \\
$k = 0.4$ & 22.52 / 22.53 & 28.39 / 28.43 & 24.76 / 25.24 & 26.62 / 26.87 & 28.21 / 28.15 & 28.58 / 28.77 & 23.50 / 23.77     \\
$k = 1.0$ & 22.62 / 22.48 & 28.11 / 28.71 & 24.71 / 24.70 & 26.75 / 26.61 & 27.31 / 28.15 & 28.90 / 28.91 & 23.45 / 23.62    \\
$k = 2.5$ & 22.61 / 22.52 & 28.81 / 28.40 & 24.96 / 24.96 & 26.76 / 26.67 & 27.42 / 27.72 & 28.73 / 28.50 & \textcolor{red}{18.59} / \textcolor{blue}{23.53}      \\
$k = 10.0$ & \textcolor{red}{12.42} / \textcolor{blue}{22.40} & \textcolor{red}{13.91} / \textcolor{blue}{28.50} & 24.55 / 25.26 & \textcolor{red}{14.35} / \textcolor{blue}{26.47} & 27.69 / 28.21 & \textcolor{red}{11.27} / \textcolor{blue}{28.89} & \textcolor{red}{15.91} / \textcolor{blue}{23.81}     \\
\specialrule{.15em}{.05em}{.05em}
\end{tabular}
\vspace{-0.2cm}
\caption{
PSNR values of [\textcolor{red}{baseline} / \textcolor{blue}{ours}] with different scene scalings $k$ on the Mip-NeRF 360 dataset. 
The baseline is Nerfacto, which uses the $\mathtt{exp}$ activation, and thus performs well even when $k$ is small. Using a high transmittance initialization strategy like ours makes this model more robust at larger scene scalings.
}
\label{tab:nerfacto_360}
\vspace{-0.3cm}
\end{table*}

\subsection{Background Contraction / Disparity Sampling}
\label{sec:scene_contract}
To represent unbounded scenes, Mip-NeRF 360~\cite{mipnerf360} applies a contraction of $(2 - \frac{1}{\|x\|})\frac{x}{\|x\|} ~\text{if } \|x\|_2 > 1$ so that distant points with insufficient camera coverage use less model capacity.
Nerfstudio~\cite{nerfstudio} and MeRF~\cite{merf} modify it to be more suitable for voxel grids.
This scene contraction is part of the underlying NN blackbox that is in principle orthogonal to how we do volume rendering. What matters is the sampling strategy, which in this case is co-designed with samples spaced linearly in disparity. Importantly, metric distances (in t-space) \emph{before} the disparity transform are used as the ray interval lengths $d$ for volume rendering.
Taking Nerfacto for example, the metric ray travels from $0.05$ to $1000$, with the largest intervals $d$ close to $700$ for samples near the outer scene boundary. It provides an inductive bias that faraway background regions have alpha values close to $1.0$ at initialization.
Note that Nerfacto's ray sampler applies uniform sampling in uncontracted space and disparity sampling in contracted space. These locations are then further updated by importance sampling.
Applying scene scaling with a purely disparity based ray sampler is problematic since sample locations barely move except for the last few bins.
When applying the high transmittance offset of~\equref{eq:init}, our initial attempt is to divide the scene based on the (max) norm of sampled location. If $\|x\|_{\infty} \le 1$ we calculate the offset with $L_{\text{in}} = 2.0$ as usual. If $\|x\|_{\infty} > 1$, we try using $L_{\text{out}} = 0.0$, $2.0$, $4.0$ or $1000.0$. $L_{\text{out}} = 1000.0$ would undo the background inductive bias and in practice results in floaters in near-camera regions.
Not applying any transmittance offset at all i.e. $L_{\text{out}} = 0$ is not good either when scene scaling factor $k$ is large. See~\tabref{tab:scene_contraction} and ~\figref{fig:scene_contraction} for comparisons of various $L_{\text{out}}$ at $k=10$. Our final recommendation is to simply let $L_{\text{out}} = L_{\text{in}}$, so that there is no need to write branching logic in the implementation for $\|x\|_{\infty} \le 1$. The same offset is therefore applied to every point in the scene.

\section{Conclusion}
We present and clarify the concept of alpha invariance. Volume density and scene size change inversely
in order to maintain identical local opacities in a radiance field. 
It is an implication of the scale-ambiguity in a 3D scene. 
We recommend parameterizing both distance and volume densities in log space, along with a high transmittance offset for robust performance across scene sizes, and verify the approach on a few commonly used architectures in the literature.

\vspace{0.15cm}
\myparagraph{Acknowledgements.} We thank PALS, 3DL and Michael Maire's lab for their advice on the manuscript. This work was supported in part by the TRI University 2.0 program.

\clearpage
{\small
\bibliographystyle{ieeenat_fullname}
\bibliography{ref}

\begin{thebibliography}{39}
\providecommand{\natexlab}[1]{#1}
\providecommand{\url}[1]{\texttt{#1}}
\expandafter\ifx\csname urlstyle\endcsname\relax
  \providecommand{\doi}[1]{doi: #1}\else
  \providecommand{\doi}{doi: \begingroup \urlstyle{rm}\Url}\fi

\bibitem[Barron et~al.(2021)Barron, Mildenhall, Tancik, Hedman, Martin-Brualla,
  and Srinivasan]{mipnerf}
Jonathan~T Barron, Ben Mildenhall, Matthew Tancik, Peter Hedman, Ricardo
  Martin-Brualla, and Pratul~P Srinivasan.
\newblock Mip-nerf: A multiscale representation for anti-aliasing neural
  radiance fields.
\newblock In \emph{Int. Conf. Comput. Vis.}, 2021.

\bibitem[Barron et~al.(2022)Barron, Mildenhall, Verbin, Srinivasan, and
  Hedman]{mipnerf360}
Jonathan~T Barron, Ben Mildenhall, Dor Verbin, Pratul~P Srinivasan, and Peter
  Hedman.
\newblock {Mip-NeRF} 360: Unbounded anti-aliased neural radiance fields.
\newblock In \emph{IEEE Conf. Comput. Vis. Pattern Recog.}, 2022.

\bibitem[Barron et~al.(2023)Barron, Mildenhall, Verbin, Srinivasan, and
  Hedman]{zipnerf}
Jonathan~T. Barron, Ben Mildenhall, Dor Verbin, Pratul~P. Srinivasan, and Peter
  Hedman.
\newblock {Zip-NeRF}: Anti-aliased grid-based neural radiance fields.
\newblock In \emph{Int. Conf. Comput. Vis.}, 2023.

\bibitem[Bishop(1995)]{bishop_book}
Christopher~M Bishop.
\newblock \emph{Neural networks for pattern recognition}.
\newblock Oxford university press, 1995.

\bibitem[Cao and Johnson(2023)]{hexplane}
Ang Cao and Justin Johnson.
\newblock Hexplane: A fast representation for dynamic scenes.
\newblock In \emph{IEEE Conf. Comput. Vis. Pattern Recog.}, 2023.

\bibitem[Chen et~al.(2022)Chen, Xu, Geiger, Yu, and Su]{tensorf}
Anpei Chen, Zexiang Xu, Andreas Geiger, Jingyi Yu, and Hao Su.
\newblock Tensorf: Tensorial radiance fields.
\newblock In \emph{Eur. Conf. Comput. Vis.}, 2022.

\bibitem[Fattal(2008)]{fattal_dehaze}
Raanan Fattal.
\newblock Single image dehazing.
\newblock \emph{ACM Trans. Graph.}, 2008.

\bibitem[Fridovich-Keil et~al.(2022)Fridovich-Keil, Yu, Tancik, Chen, Recht,
  and Kanazawa]{plenoxels}
Sara Fridovich-Keil, Alex Yu, Matthew Tancik, Qinhong Chen, Benjamin Recht, and
  Angjoo Kanazawa.
\newblock Plenoxels: Radiance fields without neural networks.
\newblock In \emph{IEEE Conf. Comput. Vis. Pattern Recog.}, 2022.

\bibitem[He et~al.(2010)He, Sun, and Tang]{kaiming_dehaze}
Kaiming He, Jian Sun, and Xiaoou Tang.
\newblock Single image haze removal using dark channel prior.
\newblock \emph{IEEE Trans. Pattern Anal. Mach. Intell.}, 2010.

\bibitem[He et~al.(2015)He, Zhang, Ren, and Sun]{he_init}
Kaiming He, Xiangyu Zhang, Shaoqing Ren, and Jian Sun.
\newblock Delving deep into rectifiers: Surpassing human-level performance on
  {ImageNet} classification, 2015.

\bibitem[Huang et~al.(2023)Huang, Qin, Zhou, Zhu, Liu, and
  Shao]{dnn_normalization}
Lei Huang, Jie Qin, Yi Zhou, Fan Zhu, Li Liu, and Ling Shao.
\newblock Normalization techniques in training dnns: Methodology, analysis and
  application.
\newblock \emph{IEEE Trans. Pattern Anal. Mach. Intell.}, 2023.

\bibitem[Kerbl et~al.(2023)Kerbl, Kopanas, Leimk{\"u}hler, and
  Drettakis]{gsplats}
Bernhard Kerbl, Georgios Kopanas, Thomas Leimk{\"u}hler, and George Drettakis.
\newblock {3D Gaussian} splatting for real-time radiance field rendering.
\newblock \emph{ACM Trans. Graph.}, 2023.

\bibitem[Keselman and Hebert(2022)]{keselman_gaussian1}
Leonid Keselman and Martial Hebert.
\newblock Approximate differentiable rendering with algebraic surfaces.
\newblock In \emph{Eur. Conf. Comput. Vis.}, 2022.

\bibitem[Keselman and Hebert(2023)]{keselman_gaussian2}
Leonid Keselman and Martial Hebert.
\newblock Flexible techniques for differentiable rendering with 3d gaussians.
\newblock \emph{arXiv preprint arXiv:2308.14737}, 2023.

\bibitem[Luiten et~al.(2023)Luiten, Kopanas, Leibe, and
  Ramanan]{luiten2023dynamic}
Jonathon Luiten, Georgios Kopanas, Bastian Leibe, and Deva Ramanan.
\newblock Dynamic 3d {Gaussians}: Tracking by persistent dynamic view
  synthesis.
\newblock \emph{arXiv preprint arXiv:2308.09713}, 2023.

\bibitem[Martin-Brualla et~al.(2021)Martin-Brualla, Radwan, Sajjadi, Barron,
  Dosovitskiy, and Duckworth]{nerfw}
Ricardo Martin-Brualla, Noha Radwan, Mehdi~SM Sajjadi, Jonathan~T Barron,
  Alexey Dosovitskiy, and Daniel Duckworth.
\newblock {NeRF} in the wild: Neural radiance fields for unconstrained photo
  collections.
\newblock In \emph{IEEE Conf. Comput. Vis. Pattern Recog.}, 2021.

\bibitem[Max(1995)]{maxvol}
Nelson Max.
\newblock Optical models for direct volume rendering.
\newblock \emph{IEEE Trans. Vis. Comput. Graph.}, 1995.

\bibitem[Meuleman et~al.(2023)Meuleman, Liu, Gao, Huang, Kim, Kim, and
  Kopf]{localrf}
Andreas Meuleman, Yu-Lun Liu, Chen Gao, Jia-Bin Huang, Changil Kim, Min~H. Kim,
  and Johannes Kopf.
\newblock Progressively optimized local radiance fields for robust view
  synthesis.
\newblock In \emph{IEEE Conf. Comput. Vis. Pattern Recog.}, 2023.

\bibitem[Mildenhall et~al.(2019)Mildenhall, Srinivasan, Ortiz-Cayon, Kalantari,
  Ramamoorthi, Ng, and Kar]{llff}
Ben Mildenhall, Pratul~P. Srinivasan, Rodrigo Ortiz-Cayon, Nima~Khademi
  Kalantari, Ravi Ramamoorthi, Ren Ng, and Abhishek Kar.
\newblock Local light field fusion: Practical view synthesis with prescriptive
  sampling guidelines.
\newblock \emph{ACM Trans. Graph.}, 2019.

\bibitem[Mildenhall et~al.(2020)Mildenhall, Srinivasan, Tancik, Barron,
  Ramamoorthi, and Ng]{nerf}
Ben Mildenhall, Pratul~P Srinivasan, Matthew Tancik, Jonathan~T Barron, Ravi
  Ramamoorthi, and Ren Ng.
\newblock Nerf: Representing scenes as neural radiance fields for view
  synthesis.
\newblock In \emph{Eur. Conf. Comput. Vis.}, pages 405--421, 2020.

\bibitem[M{\"u}ller et~al.(2022)M{\"u}ller, Evans, Schied, and
  Keller]{instantngp}
Thomas M{\"u}ller, Alex Evans, Christoph Schied, and Alexander Keller.
\newblock Instant neural graphics primitives with a multiresolution hash
  encoding.
\newblock \emph{ACM Trans. Graph.}, 2022.

\bibitem[Narasimhan and Nayar(2000)]{bad_weather2}
Srinivasa~G Narasimhan and Shree~K Nayar.
\newblock Chromatic framework for vision in bad weather.
\newblock In \emph{IEEE Conf. Comput. Vis. Pattern Recog.}, 2000.

\bibitem[Nayar and Narasimhan(1999)]{bad_weather1}
Shree~K Nayar and Srinivasa~G Narasimhan.
\newblock Vision in bad weather.
\newblock In \emph{Int. Conf. Comput. Vis.}, 1999.

\bibitem[Niemeyer et~al.(2020)Niemeyer, Mescheder, Oechsle, and Geiger]{dvr}
Michael Niemeyer, Lars Mescheder, Michael Oechsle, and Andreas Geiger.
\newblock Differentiable volumetric rendering: Learning implicit 3d
  representations without 3d supervision.
\newblock In \emph{IEEE Conf. Comput. Vis. Pattern Recog.}, 2020.

\bibitem[Penner and Zhang(2017)]{penner_zhang}
Eric Penner and Li Zhang.
\newblock Soft 3d reconstruction for view synthesis.
\newblock \emph{ACM Trans. Graph.}, 2017.

\bibitem[Pumarola et~al.(2021)Pumarola, Corona, Pons-Moll, and
  Moreno-Noguer]{dnerf}
Albert Pumarola, Enric Corona, Gerard Pons-Moll, and Francesc Moreno-Noguer.
\newblock {D-NeRF}: Neural radiance fields for dynamic scenes.
\newblock In \emph{IEEE Conf. Comput. Vis. Pattern Recog.}, 2021.

\bibitem[Reiser et~al.(2023)Reiser, Szeliski, Verbin, Srinivasan, Mildenhall,
  Geiger, Barron, and Hedman]{merf}
Christian Reiser, Rick Szeliski, Dor Verbin, Pratul Srinivasan, Ben Mildenhall,
  Andreas Geiger, Jon Barron, and Peter Hedman.
\newblock {MERF}: Memory-efficient radiance fields for real-time view synthesis
  in unbounded scenes.
\newblock \emph{ACM Trans. Graph.}, 2023.

\bibitem[Srinivasan et~al.(2019)Srinivasan, Tucker, Barron, Ramamoorthi, Ng,
  and Snavely]{srinivasan2019pushing}
Pratul~P Srinivasan, Richard Tucker, Jonathan~T Barron, Ravi Ramamoorthi, Ren
  Ng, and Noah Snavely.
\newblock Pushing the boundaries of view extrapolation with multiplane images.
\newblock In \emph{IEEE Conf. Comput. Vis. Pattern Recog.}, 2019.

\bibitem[Sun et~al.(2022)Sun, Sun, and Chen]{dvgo}
Cheng Sun, Min Sun, and Hwann-Tzong Chen.
\newblock Direct voxel grid optimization: Super-fast convergence for radiance
  fields reconstruction.
\newblock In \emph{IEEE Conf. Comput. Vis. Pattern Recog.}, 2022.

\bibitem[Tancik et~al.(2023)Tancik, Weber, Ng, Li, Yi, Kerr, Wang,
  Kristoffersen, Austin, Salahi, Ahuja, McAllister, and Kanazawa]{nerfstudio}
Matthew Tancik, Ethan Weber, Evonne Ng, Ruilong Li, Brent Yi, Justin Kerr,
  Terrance Wang, Alexander Kristoffersen, Jake Austin, Kamyar Salahi, Abhik
  Ahuja, David McAllister, and Angjoo Kanazawa.
\newblock Nerfstudio: A modular framework for neural radiance field
  development.
\newblock \emph{arXiv preprint arXiv:2302.04264}, 2023.

\bibitem[Tang(2022)]{torchngp}
Jiaxiang Tang.
\newblock Torch-ngp: a {PyTorch} implementation of instant-ngp, 2022.
\newblock https://github.com/ashawkey/torch-ngp.

\bibitem[Wang et~al.(2021)Wang, Liu, Liu, Theobalt, Komura, and Wang]{neus}
Peng Wang, Lingjie Liu, Yuan Liu, Christian Theobalt, Taku Komura, and Wenping
  Wang.
\newblock Neus: Learning neural implicit surfaces by volume rendering for
  multi-view reconstruction.
\newblock In \emph{Adv. Neural Inform. Process. Syst.}, 2021.

\bibitem[Wikipedia(2023)]{wiki:rectified_gaussian}
Wikipedia.
\newblock {Rectified Gaussian distribution} --- {W}ikipedia{,} the free
  encyclopedia.
\newblock
  \url{http://en.wikipedia.org/w/index.php?title=Rectified\%20Gaussian\%20distribution&oldid=1140140064},
  2023.

\bibitem[Wu et~al.(2023)Wu, Yi, Fang, Xie, Zhang, Wei, Liu, Tian, and
  Wang]{wu20234d}
Guanjun Wu, Taoran Yi, Jiemin Fang, Lingxi Xie, Xiaopeng Zhang, Wei Wei, Wenyu
  Liu, Qi Tian, and Xinggang Wang.
\newblock 4d {Gaussian} splatting for real-time dynamic scene rendering.
\newblock \emph{arXiv preprint arXiv:2310.08528}, 2023.

\bibitem[Yariv et~al.(2020)Yariv, Kasten, Moran, Galun, Atzmon, Ronen, and
  Lipman]{idr}
Lior Yariv, Yoni Kasten, Dror Moran, Meirav Galun, Matan Atzmon, Basri Ronen,
  and Yaron Lipman.
\newblock Multiview neural surface reconstruction by disentangling geometry and
  appearance.
\newblock In \emph{Adv. Neural Inform. Process. Syst.}, 2020.

\bibitem[Yariv et~al.(2021)Yariv, Gu, Kasten, and Lipman]{volsdf}
Lior Yariv, Jiatao Gu, Yoni Kasten, and Yaron Lipman.
\newblock Volume rendering of neural implicit surfaces.
\newblock In \emph{Adv. Neural Inform. Process. Syst.}, 2021.

\bibitem[Yen-Chen(2020)]{lin2020nerfpytorch}
Lin Yen-Chen.
\newblock {NeRF-Pytorch}.
\newblock \url{https://github.com/yenchenlin/nerf-pytorch/}, 2020.

\bibitem[Zhou et~al.(2018)Zhou, Tucker, Flynn, Fyffe, and Snavely]{zhou_mpi}
Tinghui Zhou, Richard Tucker, John Flynn, Graham Fyffe, and Noah Snavely.
\newblock Stereo magnification: Learning view synthesis using multiplane
  images.
\newblock In \emph{SIGGRAPH}, 2018.

\bibitem[Zitnick et~al.(2004)Zitnick, Kang, Uyttendaele, Winder, and
  Szeliski]{zitnick04}
C~Lawrence Zitnick, Sing~Bing Kang, Matthew Uyttendaele, Simon Winder, and
  Richard Szeliski.
\newblock High-quality video view interpolation using a layered representation.
\newblock \emph{ACM Trans. Graph.}, 2004.

\end{thebibliography}
}

\clearpage

\appendix
\newcommand{\beginsupplementary}{%
    \setcounter{section}{0}
	\renewcommand{\thesection}{A\arabic{section}}
	\renewcommand{\thesubsection}{\thesection.\arabic{subsection}}

	\renewcommand{\thetable}{A\arabic{table}}%
	\setcounter{table}{0}

	\renewcommand{\thefigure}{A\arabic{figure}}%
	\setcounter{figure}{0}

        \renewcommand{\thealgorithm}{A\arabic{algorithm}}%
	\setcounter{algorithm}{0}
}
\beginsupplementary

{\noindent\bf \LARGE Appendix
}
\vspace{0.1cm}

\section{Related Work}
\paragraph{Alpha and Bad Weather.} Early works on weather modeling~\cite{bad_weather1,bad_weather2} use volume rendering to characterize the effect of rain and fog. Dehazing methods~\cite{fattal_dehaze,kaiming_dehaze}, apart from
separating the base scene albedo from the foggy airlight, use the estimated alpha mask to compute an up-to-scale depth map proportional to $-\log(1 - \alpha)$.
The notion of alpha invariance is implied in this step.

\paragraph{NeRF vs MPI.} NeRF uses volume rendering~\cite{maxvol} for view synthesis~\cite{nerf} and other inverse rendering tasks .
Unlike prior works~\cite{zitnick04,penner_zhang,zhou_mpi,srinivasan2019pushing} that directly model the alphas of a fixed set of multi-plane images (MPI), NeRF optimizes the continuous volumetric densities $\sigma(x)$. MPI fixes the discretization in advance, whereas the density parameterization in NeRF makes it easy to adjust discretization and resample along a ray. The flipside of this flexibility, however, is that the magnitude of $\sigma(x)$ is tied to the domain of integration.
Our work emphasizes that even though $\sigma(x)$ and distance change to compensate for each other, the alpha value of a particular discretized segment should remain constant. The opacity is an invariant property of local geometry.

\paragraph{Volume-rendered SDF.} Signed distance function (SDF) is
suitable for procedural content authoring and surface extraction. SDF rendering used to rely on finding surface intersections by sphere tracing~\cite{idr,dvr}, but recent methods such as NeuS~\cite{neus} and VolSDF~\cite{volsdf} move to the “fuzzier” volume rendering for easier optimization. To perform volume rendering, the distance function is first transformed into volumetric densities before alpha compositing.
VolSDF learns scaling and shrinking coefficients on the distance-transformed volume densities.
NeuS instead demands the CDF of volume rendering to match the shape of a scaled, horizontally flipped $\mathtt{sigmoid}$ i.e. $1 - T(t) = \mathtt{sigmoid}(s \cdot -\mathtt{SDF}(t))$, so that the CDF’s derivative, in other words the weighting function $w(t)$, has its local maxima located at SDF value $0$ for precise surface level-set extraction.
The learned parameter $s$ in the $\mathtt{sigmoid}$ acts as a global scaling coefficient on the implied volume densities. In this sense both formulations are alpha invariant by default, with the extra constraint that the magnitude of volume density function is globally tied to its sharpness. The assumption is restrictive but fine for most use cases.

\paragraph{Gaussian Splatting.} Gaussian splatting~\cite{gsplats,luiten2023dynamic,wu20234d} renders an image by alpha compositing point primitives.
Since the scene representation is by nature discrete, the concept of volume density does not apply, as is the case with MPIs.
These explicit primitives can be efficiently rasterized using GPU pipelines. Whereas NeRF could use ray sampling strategies on the continuous density field to refine local details and discover missing structures,
3D gaussians / metaballs~\cite{gsplats,keselman_gaussian1,keselman_gaussian2} are less flexible; it relies on SfM initialization in some cases, and needs to explicitly optimize point shape, location, and periodically remove, repopulate and merge-split the primitives. 3DGS~\cite{gsplats} uses an exponent activation to scale the shape of each primitive, and it shares a similar motivation in terms of handling arbitrary scene scaling.

\begin{table*}[t]
\centering
\setlength{\tabcolsep}{2.4pt}
\footnotesize
\begin{tabular}{c|ccccc|ccccc|ccccc|ccccc}
\multicolumn{1}{c|}{} & \multicolumn{10}{c|}{Chair} & \multicolumn{10}{c}{Ship} \\ \hline
\multicolumn{1}{c|}{} & \multicolumn{5}{c|}{Default} & \multicolumn{5}{c|}{Ours} & \multicolumn{5}{c|}{Default} & \multicolumn{5}{c}{Ours} \\ \hline
$k = 0.1$ & 31.20 & 31.14 & 31.04 & \textcolor{red}{14.04} & \textcolor{red}{28.82} & 31.20 & 31.27 & 31.07 & 30.99 & 31.25
& \textcolor{red}{5.88} & \textcolor{red}{5.88} & 27.59 & 27.61 & \textcolor{red}{5.88} & 27.20 & 27.06 & 27.56 & 27.51 & 27.31 \\ 
$k = 0.4$ & \textcolor{red}{28.96} & 31.32 & 31.26 & 31.17 & 31.26 & 31.00 & 31.29 & 31.02 & 31.24 & 31.09
& 27.57 & 27.46 & 27.59 & 27.35 & 27.49 & 27.11 & 27.14 & 27.00 & 27.11 & 27.84 \\ 
$k = 1.0$ & \textcolor{red}{14.04} & 31.32 & \textcolor{red}{14.04} & \textcolor{red}{14.04} & \textcolor{red}{28.82} & 31.24 & 31.30 & 31.11 & 31.39 & 31.23
& 27.56 & \textcolor{red}{25.64} & 27.27 & \textcolor{red}{25.57} & 27.60 & 27.18 & 26.99 & 27.16 & 27.30 & 27.11 \\ 
$k = 2.5$ & \textcolor{red}{14.04} & 31.37 & \textcolor{red}{28.99} & 31.38 & 31.32 & 31.14 & 31.23 & 31.15 & 31.03 & 31.23
& 27.50 & 27.45 & \textcolor{red}{5.88} & 27.49 & 27.56 & 27.56 & 27.31 & 27.35 & 27.18 & 27.16 \\ 
$k = 10.0$ & \textcolor{red}{14.04} & 30.97 & \textcolor{red}{28.76} & 31.00 & \textcolor{red}{9.75} & 31.29 & 31.19 & 31.16 & 31.10 & 31.22
& \textcolor{red}{5.88} & 27.57 & 27.35 & \textcolor{red}{24.12} & 27.43 & 26.99 & 27.00 & 27.16 & 27.27 & 27.28 \\
\hline
\multicolumn{1}{c|}{average} & \multicolumn{5}{c|}{25.76 $\pm$ 7.79} & 
\multicolumn{5}{c|}{31.18 $\pm$ 0.10} & 
\multicolumn{5}{c|}{22.89 $\pm$ 8.54} & 
\multicolumn{5}{c}{27.23 $\pm$ 0.20} \\
\end{tabular}
\caption{
PSNR $\uparrow$ values of Vanilla-NeRF on the chair and ship scenes from the Blender dataset. Here, we run $5$ experiments for $5$ different $k$-values, and compare the results from the default NeRF baseline against our $\exp$ parametrization on $\sigma$ and high transmittance initialization. We observe consistent rendering quality across all runs for both scenes with our method, but identify inconsistent rendering quality for the default configuration, with failure modes and poor convergence marked here in \textcolor{red}{red}.
}
\label{tab:vanilla-nerf-appendix-reproduce}
\end{table*}

\begin{figure*}[t]
    \centering
    \includegraphics[width=\linewidth]{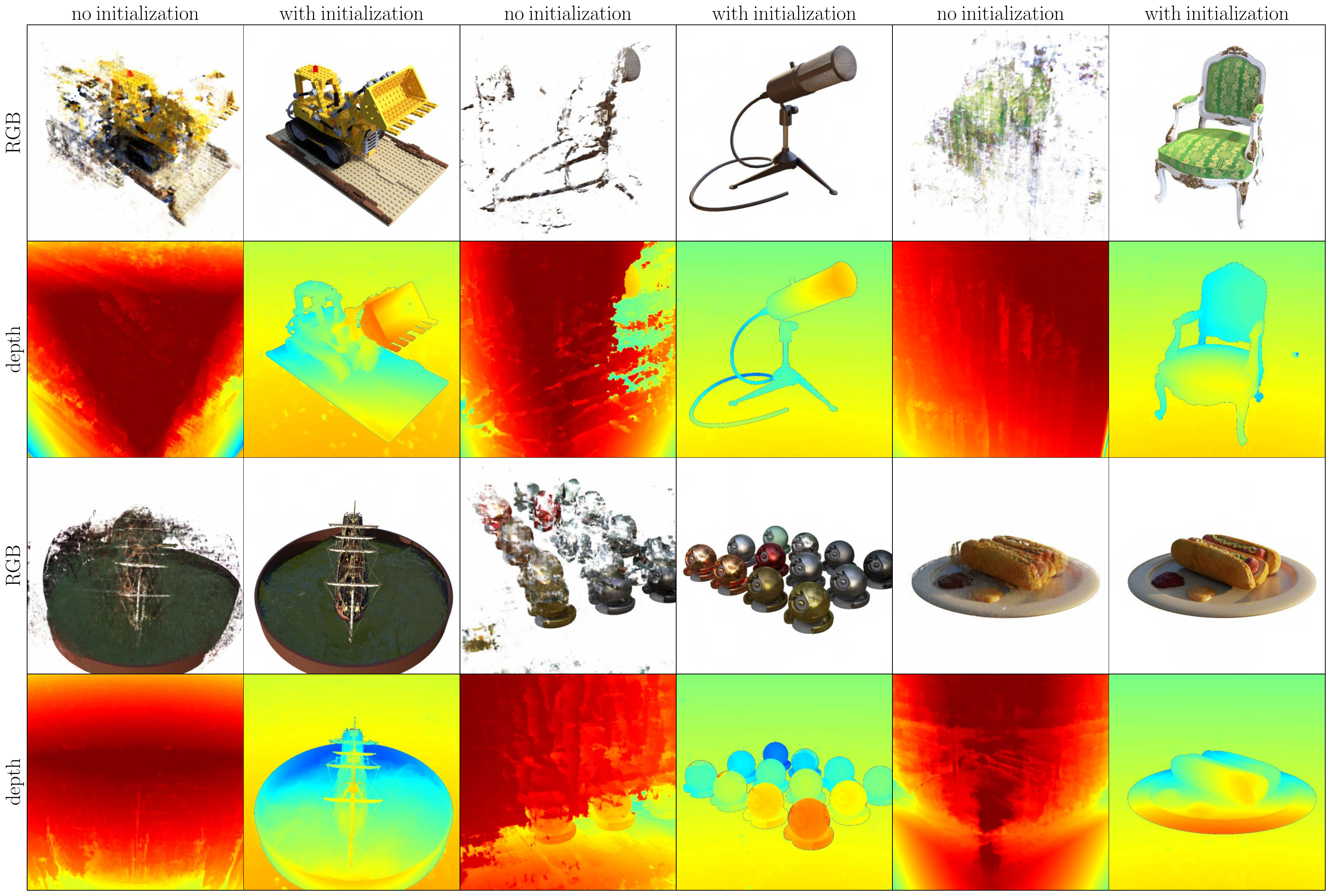}
    \caption{
     RGB-image and depth-maps produced from TensoRF on the lego, mic, chair, ship, materials, and hotdog scenes from the Blender dataset at a large scene scaling $k=25$. When using $\mathtt{exp}$ activation to parameterize $\sigma$, not using our high transmittance initialization strategy causes the optimization to get stuck with cloudy floaters.}
    \label{fig:exp_init_rgbd_tensorf_blender}
\end{figure*}

\section{Transmittance in Discrete Setting}
In \secref{sec:alphainv}, we write down the expression for high transmittance initialization in the continuous setting.
The expression is the same when the ray is cut into discrete intervals.
The tree-branching analogy in~\figref{fig:volrend} shows that the transmittance / survival probability for each segment is $1 - \alpha_i$, with overall survival probability $\prod(1 - \alpha_i) = \prod e^{-\sigma_i d_i} = e^{-\sum \sigma_i d_i} = \exp{(-\int_0^L \sigma ds)}.$  The same steps as~\equref{eq:recommend1} from \secref{sec:alphainv} follow.

\section{Additional Results}
\begin{figure*}[t]
\centering
\includegraphics[width=\linewidth]{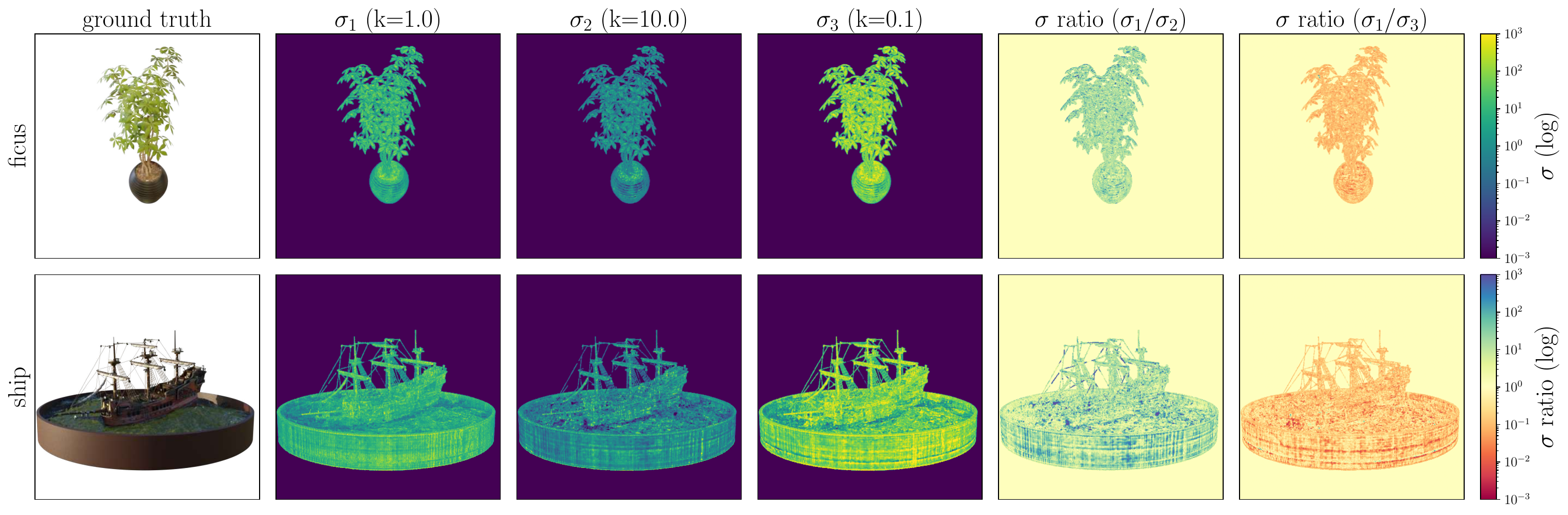}
\includegraphics[width=\linewidth]{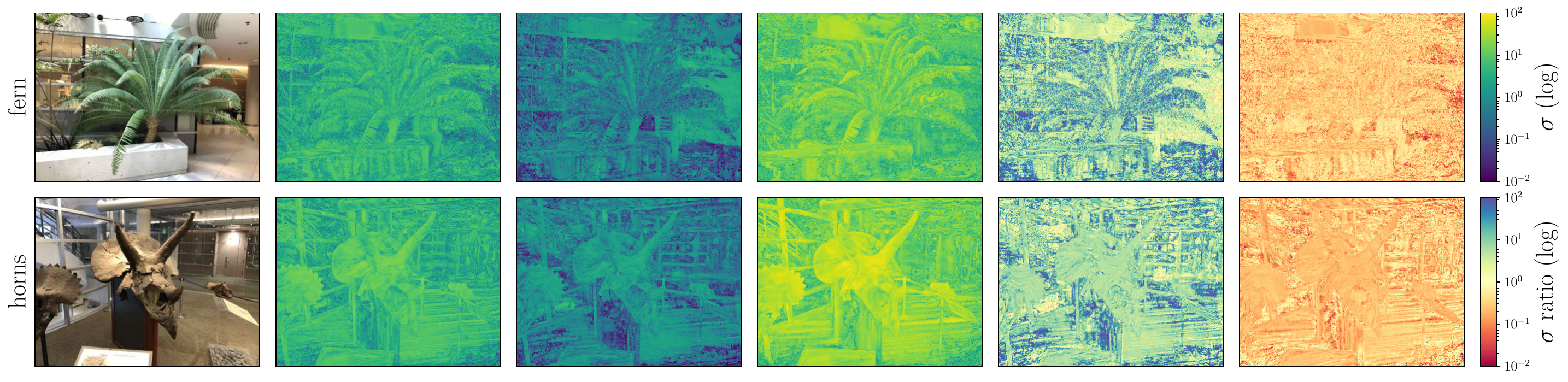}
\includegraphics[width=\linewidth]{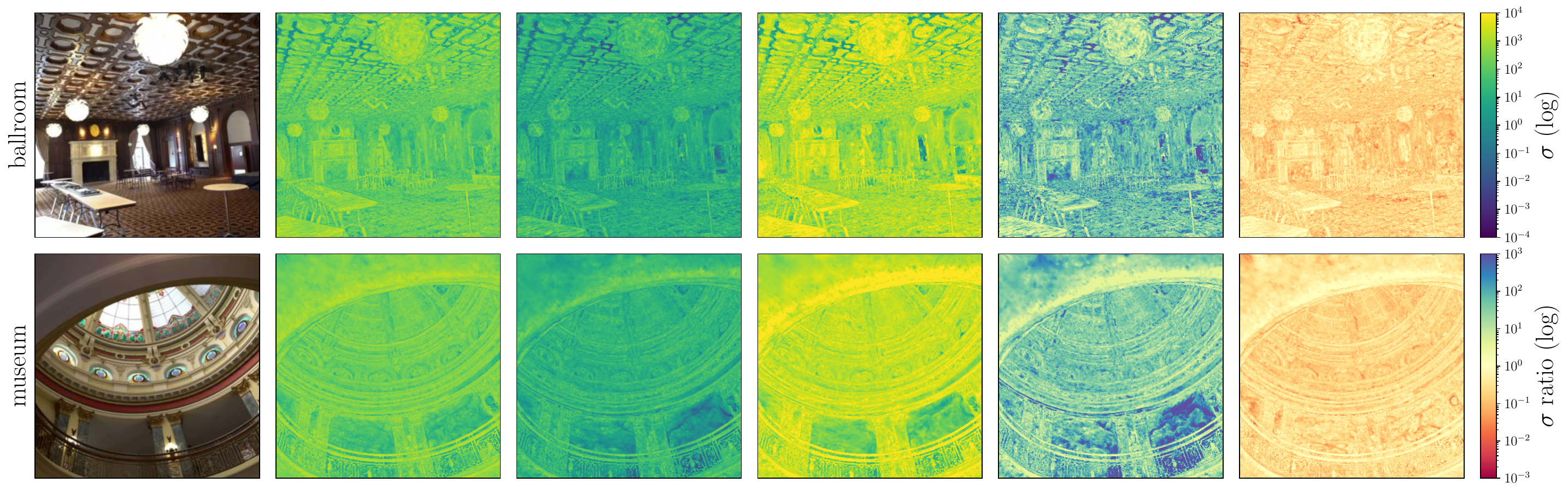}
\caption{
    Visualization of $\sigma$-image produced at the $50$-th percentile location of each ray's density histogram CDF. We produce a division image that shows the global difference of numerical range across different scene scaling factor $k$. The first four rows are produced from the TensoRF architecture (the top two rows are the ficus and ship scenes from the blender dataset; the middle two rows are the fern and horns scenes from the LLFF dataset). The bottom two rows are produced from the Nerfacto architecture on the ballroom and museum scenes from the Tanks-and-Temples dataset. Across a variety of scenes and NeRF architectures, these visualizations demonstrate that the phenomenon of alpha invariance holds to a very strong degree.%
}
\label{fig:appendix50_percentile}
\vspace{-1em}
\end{figure*}

\myparagraph{Surface Statistics.} We provide additional visualizations verifying alpha invariance in~\figref{fig:appendix50_percentile}. Similar to~\figref{fig:all_50_percentile_sigma_viz}, we shoot rays through each pixel to obtain the density histograms $\{w_i\}$, and query the spatial location at the $50$-th percentile of the probability CDF of each histogram for its $\sigma$ value. Here, we showcase various scene types from different datasets, and generate these visualizations with different architectures, demonstrating how inverse scaling between distance and volume density is a generalizable phenomenon in radiance fields.

\myparagraph{Voxel Variants, continued.}
DVGO has a strategy of fixing the scene size to some canonical scale where the interval length between each sampled point is dependent on the current voxel grid resolution's ratio to the base resolution, which empirically equates to either $0.5$ or $1$, depending on the stages of optimization progress.
We also note that DVGO's sampling procedure is stochastic, as each ray has a potentially unique number of samples.
The implication is that every ray will have a different ray length purely determined by its number of samples as the interval length between any two contiguous samples is hardcoded to either $0.5$ or $1$. As such, the term ``ray length'' loses its meaning in DVGO's context as there are no deterministic near and far planes; every ray is simply deconstructed into its constituent samples.
We observe that disabling this heuristic (\ie, setting the interval lengths to be the true physical distances and forcing each ray to be bounded within a predetermined global near and far plane) produces significantly worsened rendering quality, as shown in~\tabref{tab:dvgo_all}.

For Plenoxels, only rectifying $\sigma$ with $\exp$ is insufficient to maintain convergence at various scene scales; a high transmittance offset is needed even at $k=1$. See~\tabref{tab:plenoxels_appendix} for numerical results.
We note that our results are worse ($\sim$ 2-3 dB) than the results presented in Plenoxels~\cite{plenoxels}.
We had to significantly lower the learning rate to be more suitable for the $\exp$ activation, and believe that other changes in the hyperparameters are also needed to match the default performance.
We leave this as future work; our results still demonstrate consistent rendering quality and a need for our high transmittance initialization strategy.

\myparagraph{Benefit of High Transmittance Initialization.} We provide additional RGB-image and depth-map visuals in~\figref{fig:exp_init_rgbd_tensorf_blender} demonstrating the benefits of our high transmittance initialization in handling large scene scales when using $\exp$ activation.

\begin{table*}[t]
\centering
\setlength{\tabcolsep}{3.9pt}
\small
\begin{tabular}{ccccccccc}
\specialrule{.15em}{.05em}{.05em}
 & \textbf{chair} & \textbf{drums} & \textbf{ficus} & \textbf{hotdog} & \textbf{lego} & \textbf{materials} & \textbf{mic} & \textbf{ship} \\
\midrule
$\mathrm{disabled}$ & \textcolor{red}{fail} & \textcolor{red}{fail} & \textcolor{red}{fail} & \textcolor{red}{fail} & \textcolor{red}{fail} & \textcolor{red}{fail} & \textcolor{red}{fail} & \textcolor{red}{fail} \\

$\mathrm{default}$ & 34.10 & 25.40 & 32.56 & 36.67 & 34.53 & 29.71 & 33.23 & 28.76   \\
\end{tabular}
\begin{tabular}{ccccccccc}
\specialrule{.15em}{.05em}{.05em}
 & \textbf{fern} & \textbf{flower} & \textbf{fortress} & \textbf{horns} & \textbf{leaves} & \textbf{orchids} & \textbf{room} & \textbf{trex} \\ 
\midrule
$\mathrm{disabled}$ & \textcolor{red}{15.74} & \textcolor{red}{17.38} & \textcolor{red}{20.77} & \textcolor{red}{20.62} & \textcolor{red}{16.96} & \textcolor{red}{11.77} & \textcolor{red}{21.44} & \textcolor{red}{20.62} \\
$\mathrm{default}$ & 24.49 & 27.61 & 29.91 & 27.01 & 20.41 & 19.95 & 30.87 & 26.41  \\
\end{tabular}
\begin{tabular}{cccccccc}
\specialrule{.15em}{.05em}{.05em}
 & \textbf{bicycle} & \textbf{bonsai} & \textbf{counter} & \textbf{garden} & \textbf{kitchen} & \textbf{room} & \textbf{stump} \\
\midrule
$\mathrm{disabled}$ & \textcolor{red}{fail} & \textcolor{red}{fail} & \textcolor{red}{fail} & \textcolor{red}{fail} & \textcolor{red}{fail} & \textcolor{red}{fail} & \textcolor{red}{fail} \\
$\mathrm{default}$ & 21.98 & 27.33 & 25.41 & 24.41 & 25.81 & 28.14 & 23.51  \\
\specialrule{.15em}{.05em}{.05em}
\end{tabular}
\caption{
PSNR $\uparrow$ values of DVGO on the Blender (top row), LLFF (middle row), and Mip-NeRF 360 (bottom row) datasets. 
DVGO sets the interval lengths $d$ to the ratio of the current voxel grid resolution to the base voxel grid resolution in order to to make the model independent of scene size; this is referred to as `$\mathrm{default}$' in the table.
We observe that disabling this heuristic (i.e., setting the interval lengths in the volume rendering equation to be the true physical distances between any two sample points) results in DVGO failing to render at a high quality. These failure modes are marked with \textcolor{red}{red} in the rows titled `$\mathrm{disabled}$'.
}
\label{tab:dvgo_all}
\end{table*}

\begin{table*}[t]
\centering
\setlength{\tabcolsep}{3.9pt}
\small
\begin{tabular}{lcccccccc}
\specialrule{.15em}{.05em}{.05em}
 & \textbf{chair} & \textbf{drums} & \textbf{ficus} & \textbf{hotdog} & \textbf{lego} & \textbf{materials} & \textbf{mic} & \textbf{ship} \\
\midrule
$k = 0.1$ & 31.80 / 31.89 & 24.23 / 24.26 & 29.41 / 29.58 & 34.12 / 34.31 & 31.20 / 31.31 & 28.21 / 28.41 & 31.11 / 31.31 & 28.16 / 28.25  \\
$k = 0.4$ & \textcolor{red}{30.33} / \textcolor{blue}{31.83} & 
23.61 / 24.23 & 
\textcolor{red}{27.65} / \textcolor{blue}{29.12} & 
\textcolor{red}{31.76} / \textcolor{blue}{34.46} & 
\textcolor{red}{29.61} / \textcolor{blue}{31.22} & 
\textcolor{red}{26.06} / \textcolor{blue}{28.02} & 
\textcolor{red}{29.04} / \textcolor{blue}{31.13} & 
\textcolor{red}{27.01} / \textcolor{blue}{28.13} \\
$k = 1.0$ & \textcolor{red}{27.06} / \textcolor{blue}{32.00} & 
\textcolor{red}{21.66} / \textcolor{blue}{24.37} & 
\textcolor{red}{24.77} / \textcolor{blue}{29.39} & 
\textcolor{red}{27.31} / \textcolor{blue}{34.56} & 
\textcolor{red}{26.22} / \textcolor{blue}{31.35} & 
\textcolor{red}{23.14} / \textcolor{blue}{28.31} & 
\textcolor{red}{24.85} / \textcolor{blue}{31.29} & 
\textcolor{red}{24.58} / \textcolor{blue}{28.19} \\
$k = 2.5$ & \textcolor{red}{17.49} / \textcolor{blue}{32.43} & 
\textcolor{red}{17.24} / \textcolor{blue}{24.44} & 
\textcolor{red}{18.89} / \textcolor{blue}{29.83} & 
\textcolor{red}{20.20} / \textcolor{blue}{34.72} & 
\textcolor{red}{16.64} / \textcolor{blue}{31.53} & 
\textcolor{red}{15.51} / \textcolor{blue}{28.64} & 
\textcolor{red}{15.89} / \textcolor{blue}{31.61} & 
\textcolor{red}{16.67} / \textcolor{blue}{28.22} \\
$k = 10.0$ & \textcolor{red}{13.05} / \textcolor{blue}{32.25} & 
\textcolor{red}{10.84} / \textcolor{blue}{24.45} & 
\textcolor{red}{11.40} / \textcolor{blue}{30.23} & 
\textcolor{red}{13.03} / \textcolor{blue}{34.93} & 
\textcolor{red}{11.64} / \textcolor{blue}{31.69} & 
\textcolor{red}{9.57} / \textcolor{blue}{28.79} & 
\textcolor{red}{15.89} / \textcolor{blue}{31.61} & 
\textcolor{red}{11.27} / \textcolor{blue}{28.21} \\
\specialrule{.15em}{.05em}{.05em}
\end{tabular}
\vspace{-0.25cm}
\caption{
PSNR $\uparrow$ for Plenoxels [\textcolor{red}{baseline} / \textcolor{blue}{ours}] at different scene scaling $k$ on the Blender dataset. 
By default, Plenoxels applies a very large learning rate directly on the coefficients of a voxel grid, where $\sigma$ is queried via trilinear interpolation, followed by $\mathrm{ReLU}$ to ensure non-negative density values.
The baseline Plenoxels model here replaces $\mathrm{ReLU}$ with $\mathtt{exp}$ activation and uses a reduced learning rate ($\eta_{\mathrm{init}} = 0.05$, $\eta_{\mathrm{final}} = 0.005$) with no reverse cosine delay.
Our model in addition applies a high transmittance offset. 
As shown, this transmittance offset is required for high rendering quality at various scene sizes, even at $k=1$.
}
\label{tab:plenoxels_appendix}
\vspace{-0.25cm}
\end{table*}

\section{Reproducibility}
\label{sec:repro}
In \tabref{tab:vanilla-nerf-blender}, we observe random failure modes when training the vanilla 8-layer MLP on various scenes using NeRF-Pytorch~\cite{lin2020nerfpytorch} at git commit hash \texttt{63a5a63}. To get a better sense of the \emph{frequency} of these failure modes, we train Vanilla-NeRF on chair and ship scenes $5$ times for each $k$-value, comparing the default $\mathtt{ReLU}$ parametrization on $\sigma$ against our high transmittance initialization in combination with $\exp$.
Full results are provided in~\tabref{tab:vanilla-nerf-appendix-reproduce}.
We attribute the random failure modes to poor initialization of the MLP layers (the initial output distribution has $0$ mean and $0$ variance) and the inability of $\mathtt{ReLU}$ to provide a smooth transition from low to high $\alpha$ values. See~\figref{fig:act} for a visualization.
However, even with such a poorly initialized MLP, our parametrization on $\sigma$ is enough to guarantee convergence on all runs across all tested $k$ values.

We train NeRF-Pytorch for only 200k iterations. A longer training schedule of 500k iterations would push the NeRF-Pytorch performance closer, but still slightly below the original NeRF numbers. The overall conclusions do not change.

\end{document}